\documentclass[11pt]{article}

\usepackage[final]{acl}

\usepackage{times}
\usepackage{latexsym}
\usepackage{tcolorbox}

\usepackage[T1]{fontenc}

\usepackage[utf8]{inputenc}

\usepackage{microtype}

\usepackage{inconsolata}

\usepackage{graphicx}
\usepackage{booktabs}
\usepackage[dvipsnames]{xcolor}
\usepackage[table]{xcolor}
\usepackage{multirow}
\usepackage{makecell}

\newcommand{\posdelta}[2]{\cellcolor{green!#2}\textcolor{green!45!black}{#1}}
\newcommand{\negdelta}[2]{\cellcolor{red!#2}\textcolor{red!55!black}{#1}}
\newcommand{\zerodelta}{\cellcolor{gray!6}{0.00}}

\usepackage{comment}

\title{Slow to See, Slow to Suppress: Understanding the Effects of Modality in Context-Memory Conflicts}

\author{Athulith Paraselli \\
  Brown University \\
  \texttt{athulith\_paraselli@brown.edu} \\\And
  Etha Tianze Hua \\
  Brown University \\
  \texttt{tianze\_hua@brown.edu} \\ \\\And
  Ellie Pavlick \\
  Brown University \\
  \texttt{ellie\_pavlick@brown.edu} \\
  }

\begin{document}
\maketitle
\begin{abstract}
 
We investigate how vision-language models (VLMs) handle context-memory conflicts; that is, situations in which the model is given information in context that differs from what was stored parametrically during training. We document asymmetric biases: models tend to prefer in-context information about entities which appear in text, but prefer parametric information about entities which appear in images. We relate this asymmetry to the late representational alignment across modalities, showing that the longer processing time associated with resolving visual entities prevents the suppression of the model's usual factual recall mechanism, thus resulting in more parametric answers. Chain-of-thought reasoning does not appear to resolve the gap, but increasing the amount of visual information in the context does show an effect. These results illustrate the complexity of ensuring consistent behavior as models become increasingly multimodal and retrieval-augmented.\footnote{Code and dataset are available at \url{https://github.com/aparaselli/slow-to-see-slow-to-suppress}}

\end{abstract}

\section{Introduction}

Modern AI systems are often expected to rely on contextual information when such information is more current, task-specific, or factually reliable than the information encoded in their parametric weights. This routinely happens, for example, as frontier systems integrate retrieval-augmented generation (RAG) \citep{DBLP:journals/corr/abs-2312-10997} or other types of tools \cite{schick2023toolformer} in order to provide information to the model that is likely to be unavailable or unreliable in the model's weights.

Prior work has investigated how large language models (LLMs) respond to conflicts between in-context and parametric knowledge. For example, \citet{yu-etal-2023-characterizing} characterize specific mechanisms that mediate models' tendency to recall parametric vs.\ in-context information in toy settings, while \citet{xie2024adaptive} and \citet{kortukov2024studying} describe models' biases in more realistic scenarios. Increasingly, however, retrieval augmentation and tool use are not restricted to text, but can include diverse modalities and data types \citep{abootorabi2025ask}. Little is known about how models recognize and reconcile these conflicts when they cross the boundary of modalities. 

In this work, we investigate context-memory conflicts in vision-language models (VLMs). We use a controlled fact-retrieval setting in which in-context information includes a mix of text and image information, designed to emulate contexts that might result from multimodal RAG pipelines. First, we demonstrate asymmetric biases across modalities: VLMs tend to prefer in-context information for entities mentioned in text, but prefer parametric information for entities which appear in images (\S\ref{sec:behavioral-asymmetry}). Then, we identify a mechanism for conflict resolution which explains this asymmetry (\S\ref{sec:mechanism}). Specifically, we find that: in the case of textual entities, VLMs rely on attention to detect the potential conflict and accordingly suppress the early-layer MLPs, which are usually responsible for parametric information retrieval \citep{geva2023dissecting}; in the case of image entities, however, VLMs are slower to resolve the in-context reference \cite{toolatetorecall} and thus fail to suppress the relevant MLPs, leading models to produce the parametric answer. Finally, we consider several prompt-based approaches to mitigate this asymmetry (\S\ref{sec:mitigation}). We find that chain-of-thought (CoT) has little effect on mitigating the modality asymmetry, but that the presence of more image information in context reduces the parametric bias for visual entities. Taken together, our results point towards important mechanisms which influence our ability to reliably update the information provided by multimodal, retrieval-augmented AI systems. In summary, we make the following contributions:
\begin{enumerate}
    \item We curate a conflicting fact-retrieval dataset with 37K instances across three domains.

    \item We show that, under text-only context, VLMs exhibit an asymmetric behavioral tendency: they prefer in-context information when an entity is presented textually, but prefer parametric information when the same entity is presented visually.

    \item We identify an entity-resolution-conditioned factual-recall suppression mechanism that explains this asymmetry. 

    \item We evaluate prompt-based interventions for reducing this asymmetry. We find that CoT prompting has little effect, while increasing the amount of image-grounded contextual information reduces the parametric bias for visual entities.
\end{enumerate}

\begin{figure}[t]
  \includegraphics[width=\columnwidth]{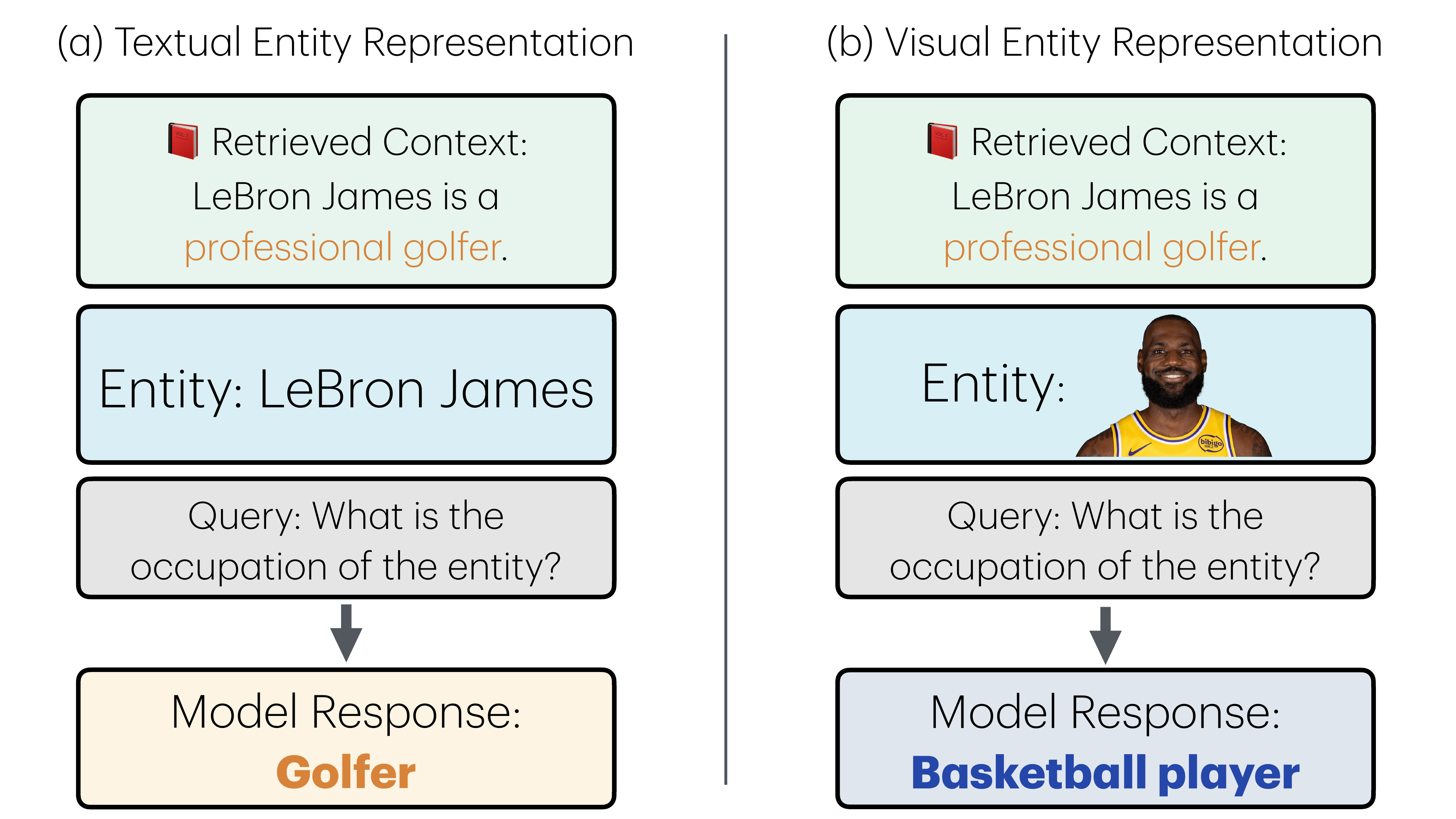}
  \caption{Context-memory conflict prompted with a textual entity (a) and a visual entity (b).}
  \label{fig:exp_setup}
\end{figure}
\section{Related work}

\paragraph{Knowledge Retrieval in LLM/VLMs.}
Knowledge retrieval in LLMs relies on a two-step process: the MLPs at the entity positions enrich entity representations sufficiently such that the attention heads at the generation tokens can extract the desired relation and promote the final answer(\citealp{meng2022locating}; \citealp{chughtai2024summing}; \citealp{geva2023dissecting}).
Recent work has compared knowledge retrieval in VLMs in an attempt to understand inconsistencies between the two. Specifically, \citet{cohen2025performancegap} showed that VLMs exhibit performance degradation on factual recall tasks when given visual inputs compared to when given textual inputs. To understand this phenomenon, recent work provided evidence that this gap is caused by a two-hop problem, where failure to accurately recall a fact is due to the entity resolution of the visual input happening too late to utilize MLPs responsible for factual recall \citep{toolatetorecall}. Additionally, \citet{sametaskdiffcircuit} discovered that VLMs use disjoint circuits to complete factual recall depending on the modality of the entity. 

\paragraph{Context-Memory Conflicts.}
A variety of benchmarks have been released to test model behavior when presented with conflicting evidence regarding a visual entity \citep{du2025mmke, Jia_Du_Jiang_Liang_Ren_Xin_Yang_Feng_Chen_Lu_etal._2026,liu-etal-2025-insight}. However, these studies focus on evaluating performance discrepancies across different VLM architectures, leaving the direct, controlled comparison between textual and visual entities unexamined. Another line of work has explored how context-memory conflicts are resolved in LLMs, identifying specific attention heads that influence final outputs and noting a correlation between training data frequency and the tendency to provide parametric answers (\citealp{kortukov2024studying}; \citealp{xie2024adaptive}). More specifically, \citet{jin-etal-2024-cutting} identify distinct attention heads which retrieve either in-context or parametric information, and show that pruning these heads can shift the model's preference between the two sources. Furthermore, \citet{li2025taming} and \citet{zhao-etal-2025-steering} utilize these insights to propose methods capable of steering a model to answer either more contextual or parametric. More recently, \citet{an2026enhancing} extend conflict mitigation techniques to VLMs, by pruning neurons which encode parametric knowledge.

\begin{table}[t]
\centering
\small
\setlength{\tabcolsep}{4pt}
\begin{tabular}{@{}llrl@{}}
\toprule
\textbf{Domain} & \textbf{Question types} & \textbf{\# Pairs} & \textbf{Sources} \\
\midrule
Celebrities & \makecell[l]{Occupation\\Birth year\\Nationality} & 2,856 & MMKC-Bench, WC \\
\midrule
Buildings   & \makecell[l]{Location\\Completion year\\Architect} & 16,230 & GLDv2, WC \\
\midrule
Artwork     & \makecell[l]{Artist\\Current location} & 18,884 & WC \\
\midrule
\textbf{Total} & & \textbf{37,970} & \\
\bottomrule
\end{tabular}
\caption{Composition of our counterfactual context benchmark. Each pair consists of an entity paired with a context statement that contradicts a known fact about it, enabling measurement of context-memory conflict resolution across modalities.}
\label{tab:benchmark}
\end{table}
\section{Experimental Setup}
\label{sec:experimental-design}

To study context-memory conflicts across modalities, we need data samples in which the same entity can be introduced either textually or visually, while the model is asked to answer a factoid question whose in-context answer differs from the answer likely stored in the model's parametric knowledge. This setup consists of three components: (1) entities with both textual names and images, (2) factoid questions with known parametric answers, and (3) controlled in-context statements that provide plausible but counterfactual alternatives to those answers.

To this end, we curate a benchmark of contradicting factoid pairs across three domains: celebrities, buildings, and artwork. The celebrity subset comprises 2,856 pairs derived from MMKC-bench \citep{Jia_Du_Jiang_Liang_Ren_Xin_Yang_Feng_Chen_Lu_etal._2026}, supplemented with additional images and facts from Wikimedia Commons \citep{WikimediaCommons2026}. Each question in this subset belongs to one of three categories: occupation, birth year, or nationality. For the architectural domain, we compiled 16,230 pairs using images from the Google Landmarks Dataset v2 \citep{weyand2020GLDv2}. These questions address location, completion year, and architect. Finally, we developed 18,884 artwork pairs sourced from Wikimedia Commons, focusing on the artist and current location. The composition of our benchmark is summarized in Table~\ref{tab:benchmark}.

\begin{table}[!t]
\centering
\small
\setlength{\tabcolsep}{4.0pt}
\renewcommand{\arraystretch}{1}
\resizebox{\columnwidth}{!}{%
\begin{tabular}{lllccc}
\toprule
\textbf{Model} & \textbf{Dataset} & \textbf{Relation} & \textbf{Text} & \textbf{Vision} & \textbf{$\Delta$ (95\% CI)} \\
\midrule
\multirow{8}{*}{\rotatebox{90}{Gemma-3-12B}} & \multirow{3}{*}{Celebrity} & \underline{Occupation} & 9.65 & \textbf{24.56} & \posdelta{+14.91\ {\scriptsize(10.09, 19.74)}}{8} \\
 &  & Birth year & 2.94 & 2.94 & \zerodelta{{\scriptsize(0.00, 0.00)}} \\
 &  & Nationality & 3.97 & \textbf{14.68} & \posdelta{+10.71\ {\scriptsize(6.74, 15.08)}}{6} \\
\cmidrule(lr){2-6}
 & \multirow{3}{*}{Building} & \underline{Country loc.} & 2.23 & \textbf{83.04} & \posdelta{+80.80\ {\scriptsize(75.45, 85.71)}}{45} \\
 &  & Built yr. & 0.00 & \textbf{1.56} & \posdelta{+1.56\ {\scriptsize(0.00, 4.69)}}{1} \\
 &  & Architect & 0.00 & 0.00 & \zerodelta{{\scriptsize(0.00, 0.00)}} \\
\cmidrule(lr){2-6}
 & \multirow{2}{*}{Artwork} & \underline{Artist name} & 5.56 & \textbf{37.41} & \posdelta{+31.85\ {\scriptsize(26.30, 37.41)}}{18} \\
 &  & Current loc. & \textbf{4.72} & 4.33 & \negdelta{-0.39\ {\scriptsize(-3.15, 2.37)}}{0} \\
\midrule
\multirow{8}{*}{\rotatebox{90}{Gemma-3-27B}} & \multirow{3}{*}{Celebrity} & \underline{Occupation} & 11.57 & \textbf{13.22} & \posdelta{+1.65\ {\scriptsize(0.00, 3.72)}}{1} \\
 &  & Birth year & 1.49 & 1.49 & \zerodelta{{\scriptsize(0.00, 0.00)}} \\
 &  & Nationality & \textbf{1.64} & 1.23 & \negdelta{-0.41\ {\scriptsize(-1.23, 0.00)}}{0} \\
\cmidrule(lr){2-6}
 & \multirow{3}{*}{Building} & \underline{Country loc.} & \textbf{3.59} & 2.94 & \negdelta{-0.65\ {\scriptsize(-1.63, 0.00)}}{0} \\
 &  & Built yr. & 0.00 & 0.00 & \zerodelta{{\scriptsize(0.00, 0.00)}} \\
 &  & Architect & 0.00 & 0.00 & \zerodelta{{\scriptsize(0.00, 0.00)}} \\
\cmidrule(lr){2-6}
 & \multirow{2}{*}{Artwork} & \underline{Artist name} & \textbf{13.10} & 6.67 & \negdelta{-6.43\ {\scriptsize(-9.76, -3.33)}}{4} \\
 &  & Current loc. & \textbf{2.82} & 1.13 & \negdelta{-1.69\ {\scriptsize(-3.11, -0.56)}}{1} \\
\midrule
\multirow{8}{*}{\rotatebox{90}{Ministral-3-8B}} & \multirow{3}{*}{Celebrity} & \underline{Occupation} & 48.91 & \textbf{77.17} & \posdelta{+28.26\ {\scriptsize(21.18, 35.87)}}{16} \\
 &  & Birth year & 0.00 & 0.00 & \zerodelta{{\scriptsize(0.00, 0.00)}} \\
 &  & Nationality & 49.38 & \textbf{70.37} & \posdelta{+20.99\ {\scriptsize(14.81, 27.78)}}{12} \\
\cmidrule(lr){2-6}
 & \multirow{3}{*}{Building} & Country loc. & 88.76 & \textbf{98.88} & \posdelta{+10.11\ {\scriptsize(6.18, 14.61)}}{6} \\
 &  & Built yr. & 56.25 & \textbf{68.75} & \posdelta{+12.50\ {\scriptsize(3.12, 25.00)}}{7} \\
 &  & \underline{Architect} & 30.00 & \textbf{60.00} & \posdelta{+30.00\ {\scriptsize(0.00, 60.00)}}{17} \\
\cmidrule(lr){2-6}
 & \multirow{2}{*}{Artwork} & \underline{Artist name} & 25.93 & \textbf{74.54} & \posdelta{+48.61\ {\scriptsize(42.13, 55.10)}}{27} \\
 &  & Current loc. & 8.09 & \textbf{17.28} & \posdelta{+9.19\ {\scriptsize(5.51, 13.24)}}{5} \\
\midrule
\multirow{8}{*}{\rotatebox{90}{Ministral-3-14B}} & \multirow{3}{*}{Celebrity} & Occupation & 59.80 & \textbf{74.51} & \posdelta{+14.71\ {\scriptsize(8.82, 21.08)}}{8} \\
 &  & Birth year & \textbf{7.00} & 4.00 & \negdelta{-3.00\ {\scriptsize(-8.00, 1.00)}}{2} \\
 &  & \underline{Nationality} & 56.06 & \textbf{77.27} & \posdelta{+21.21\ {\scriptsize(15.15, 27.78)}}{12} \\
\cmidrule(lr){2-6}
 & \multirow{3}{*}{Building} & Country loc. & 91.11 & \textbf{98.33} & \posdelta{+7.22\ {\scriptsize(2.78, 11.67)}}{4} \\
 &  & Built yr. & 54.35 & \textbf{58.70} & \posdelta{+4.35\ {\scriptsize(-6.52, 17.39)}}{2} \\
 &  & \underline{Architect} & 13.16 & \textbf{47.37} & \posdelta{+34.21\ {\scriptsize(21.05, 50.00)}}{19} \\
\cmidrule(lr){2-6}
 & \multirow{2}{*}{Artwork} & \underline{Artist name} & 37.40 & \textbf{84.73} & \posdelta{+47.33\ {\scriptsize(41.22, 53.44)}}{26} \\
 &  & Current loc. & 13.53 & \textbf{18.80} & \posdelta{+5.26\ {\scriptsize(1.88, 9.02)}}{3} \\
\midrule
\multirow{8}{*}{\rotatebox{90}{Qwen2.5-VL-7B}} & \multirow{3}{*}{Celebrity} & \underline{Occupation} & 25.61 & \textbf{60.67} & \posdelta{+35.06\ {\scriptsize(29.57, 40.24)}}{20} \\
 &  & Birth year & 2.11 & 2.11 & \zerodelta{{\scriptsize(0.00, 0.00)}} \\
 &  & Nationality & 26.95 & \textbf{29.55} & \posdelta{+2.60\ {\scriptsize(-0.97, 6.49)}}{1} \\
\cmidrule(lr){2-6}
 & \multirow{3}{*}{Building} & \underline{Country loc.} & 21.81 & \textbf{34.80} & \posdelta{+13.00\ {\scriptsize(8.59, 17.40)}}{7} \\
 &  & Built yr. & \textbf{2.22} & 0.00 & \negdelta{-2.22\ {\scriptsize(-5.56, 0.00)}}{1} \\
 &  & Architect & 0.00 & 0.00 & \zerodelta{{\scriptsize(0.00, 0.00)}} \\
\cmidrule(lr){2-6}
 & \multirow{2}{*}{Artwork} & \underline{Artist name} & 3.95 & \textbf{52.19} & \posdelta{+48.25\ {\scriptsize(41.67, 54.82)}}{27} \\
 &  & Current loc. & 2.99 & \textbf{7.07} & \posdelta{+4.08\ {\scriptsize(2.17, 6.25)}}{2} \\
\midrule
\multirow{8}{*}{\rotatebox{90}{Qwen2.5-VL-32B}} & \multirow{3}{*}{Celebrity} & \underline{Occupation} & 42.95 & \textbf{59.94} & \posdelta{+16.99\ {\scriptsize(11.54, 22.44)}}{9} \\
 &  & Birth year & 1.32 & \textbf{4.61} & \posdelta{+3.29\ {\scriptsize(0.66, 6.58)}}{2} \\
 &  & Nationality & \textbf{70.00} & 68.08 & \negdelta{-1.92\ {\scriptsize(-6.92, 3.08)}}{1} \\
\cmidrule(lr){2-6}
 & \multirow{3}{*}{Building} & Country loc. & 80.88 & \textbf{83.39} & \posdelta{+2.51\ {\scriptsize(-0.47, 5.49)}}{1} \\
 &  & \underline{Built yr.} & 12.88 & \textbf{32.58} & \posdelta{+19.70\ {\scriptsize(13.62, 26.52)}}{11} \\
 &  & Architect & 0.00 & \textbf{7.32} & \posdelta{+7.32\ {\scriptsize(2.44, 13.41)}}{4} \\
\cmidrule(lr){2-6}
 & \multirow{2}{*}{Artwork} & \underline{Artist name} & 24.24 & \textbf{63.64} & \posdelta{+39.39\ {\scriptsize(34.24, 44.85)}}{22} \\
 &  & Current loc. & 15.74 & \textbf{16.67} & \posdelta{+0.93\ {\scriptsize(-4.17, 6.48)}}{1} \\
\midrule
\multirow{8}{*}{\rotatebox{90}{Qwen2.5-VL-72B}} & \multirow{3}{*}{Celebrity} & \underline{Occupation} & 14.37 & \textbf{21.56} & \posdelta{+7.19\ {\scriptsize(3.59, 10.78)}}{4} \\
 &  & Birth year & 2.87 & 2.87 & \zerodelta{{\scriptsize(0.00, 0.00)}} \\
 &  & Nationality & \textbf{7.58} & 4.55 & \negdelta{-3.03\ {\scriptsize(-5.76, -0.61)}}{2} \\
\cmidrule(lr){2-6}
 & \multirow{3}{*}{Building} & \underline{Country loc.} & 35.17 & \textbf{55.51} & \posdelta{+20.34\ {\scriptsize(15.89, 24.79)}}{11} \\
 &  & Built yr. & \textbf{1.97} & 1.32 & \negdelta{-0.66\ {\scriptsize(-3.29, 1.32)}}{0} \\
 &  & Architect & 0.00 & 0.00 & \zerodelta{{\scriptsize(0.00, 0.00)}} \\
\cmidrule(lr){2-6}
 & \multirow{2}{*}{Artwork} & \underline{Artist name} & 14.52 & \textbf{24.45} & \posdelta{+9.93\ {\scriptsize(6.80, 13.05)}}{6} \\
 &  & Current loc. & \textbf{5.41} & 3.61 & \negdelta{-1.80\ {\scriptsize(-3.87, 0.26)}}{1} \\
\bottomrule
\end{tabular}%
}

\caption{
Percentage of parametric answers by relation under context-memory conflict. $\Delta$ denotes Vision -- Text. The final column reports the observed modality gap together with its 95\% bootstrap confidence interval.
}
\label{tab:parametric_by_relation}
\end{table}

We evaluate modality-specific behavior using a uniform prompt template: [RAG-style Supplied Textual Context] + [Entity (Image or Text)] + [Query] (Figure~\ref{fig:exp_setup}). This symmetric structure isolates the effect of entity modality. Furthermore, we constrain the supplied context to the textual modality to reflect the default text-only augmentation nature of many RAG systems. We run the benchmark on Gemma-3-12B, Gemma-3-27B \citep{gemmateam2025gemma3technicalreport}, Qwen2.5-VL-7B, Qwen2.5-VL-32B, Qwen2.5-VL-72B \citep{bai2025qwen25vltechnicalreport}, Ministral-3-8B, and Ministral-3-14B\citep{liu2026ministral3}. To ensure reliable evaluation, we first filter the data to include only instances where the models could correctly identify the entity and recall its parametric fact in the absence of conflict. This validation step results in a final dataset consisting of over 1,000 pairs for all models (see Appendix~\ref{sec:Dataset_filtering} for filtering details).

\begin{figure*}[t]
  \includegraphics[width=\textwidth]{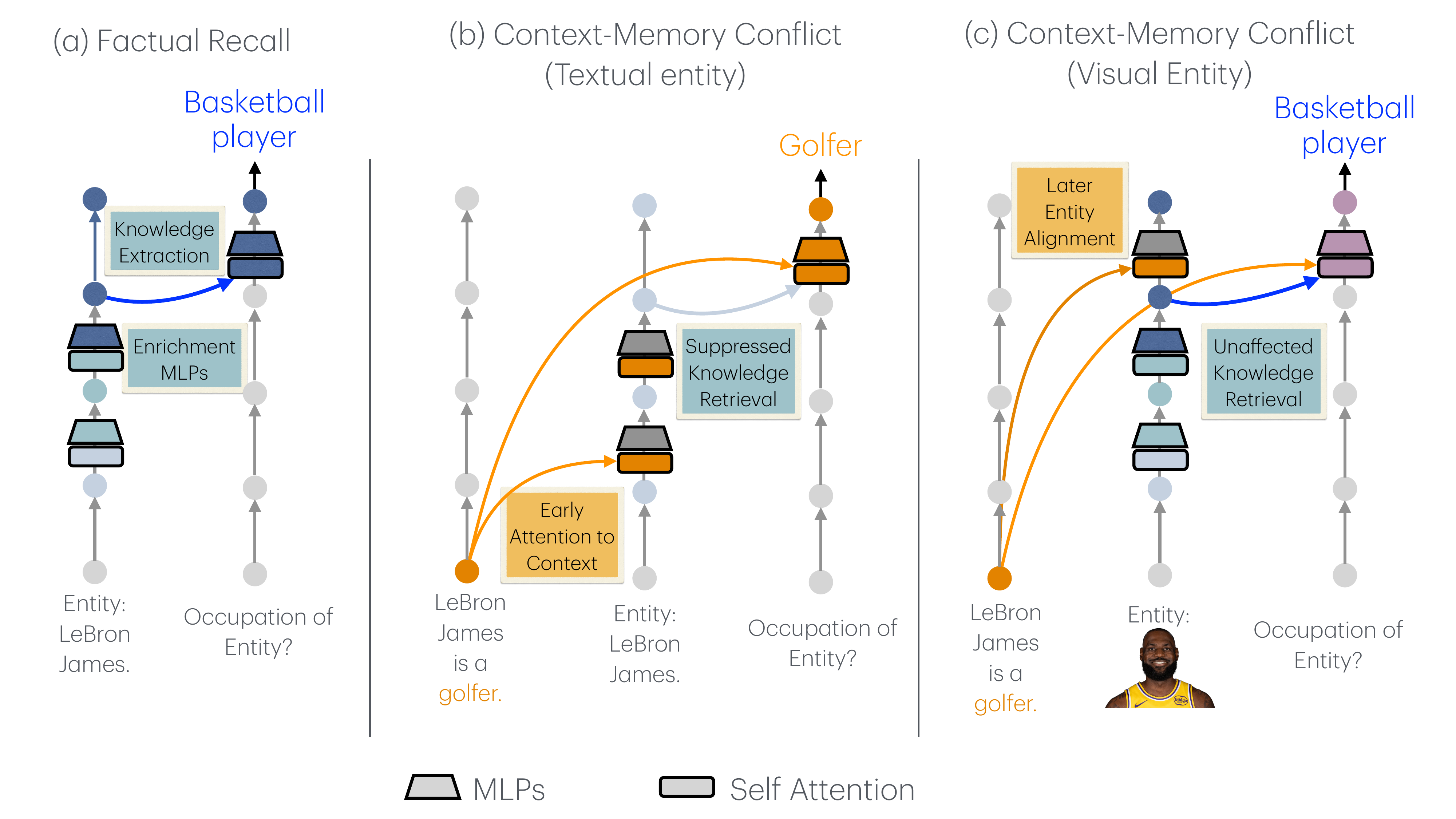}
  \caption{ Slower visual-entity resolution weakens suppression of parametric enrichment. 
(a) In standard factual recall, MLPs enrich the entity representation with parametric knowledge, enabling retrieval of associated attributes. 
(b) For textual entities under context-memory conflict, early attention to the conflicting context suppresses downstream MLP-mediated parametric enrichment, leading the model to favor the in-context answer. 
(c) For visual entities, entity resolution occurs more slowly, so this suppression arrives too late: parametric enrichment occurs largely unaffected, leading the model to favor the parametric answer.
}
  \label{fig:hypothesis}
\end{figure*}

\section{Asymmetric Conflict Resolution Between Modalities}
\label{sec:behavioral-asymmetry}

As shown in Table~\ref{tab:parametric_by_relation}, we observe that prompting VLMs with visual entities results in a substantially higher rate of parametric responses compared to their textual counterparts. This tendency is most pronounced in categories relating to essential entity traits (e.g., celebrity careers). Additionally, while most models display a parametric bias with visual prompting, the degree to which it occurs is not consistent across models and datasets. For example, building locations result in the most divergence for the Gemma-3-12B model, with an 80\% parametric bias gap. However, for Qwen2.5-VL-7B and Ministral-3-8B, the same dataset displays the least divergence. Furthermore, we observe that Ministral-3-8B has a relatively higher parametric bias for both text and visual entities. We note that Gemma-3-27B is the only tested model to show a small modality gap across all subcategories, which we explore further in Appendix~\ref{sec:gemma_mech}. However, this cannot be explained by scale alone, as the larger Qwen and Ministral models continue to display the modality gap, albeit reduced. To understand this divergent phenomenon further, we conduct a mechanistic study on the celebrity careers, building locations, and artwork artists' subcategories for the smaller model sizes of each family. We choose these subcategories as they display the largest gaps; however, in Appendix~\ref{sec:alt_rel} we explore whether the observed mechanism occurs equally in the birth-year relation, as it shows consistent behavior across modalities.

\section{Mechanism Underlying Modality Asymmetry}
\label{sec:mechanism}

We study sub-components over the entity tokens as they have previously been shown to play a crucial role in factual recall \citep{geva2023dissecting}. Through causal interventions, our findings suggest that the textual entities' attention modules detect conflicts and suppress the downstream MLPs' parametric enrichment. However, visual entities take longer to resolve \cite{toolatetorecall} and therefore do not sufficiently suppress enrichment (illustrated in Figure~\ref{fig:hypothesis}). 

\subsection{MLPs Retrieve Parametric Information}
\label{sec:mechanism:mlps}
\begin{figure*}[t]
  \includegraphics[width=\textwidth]{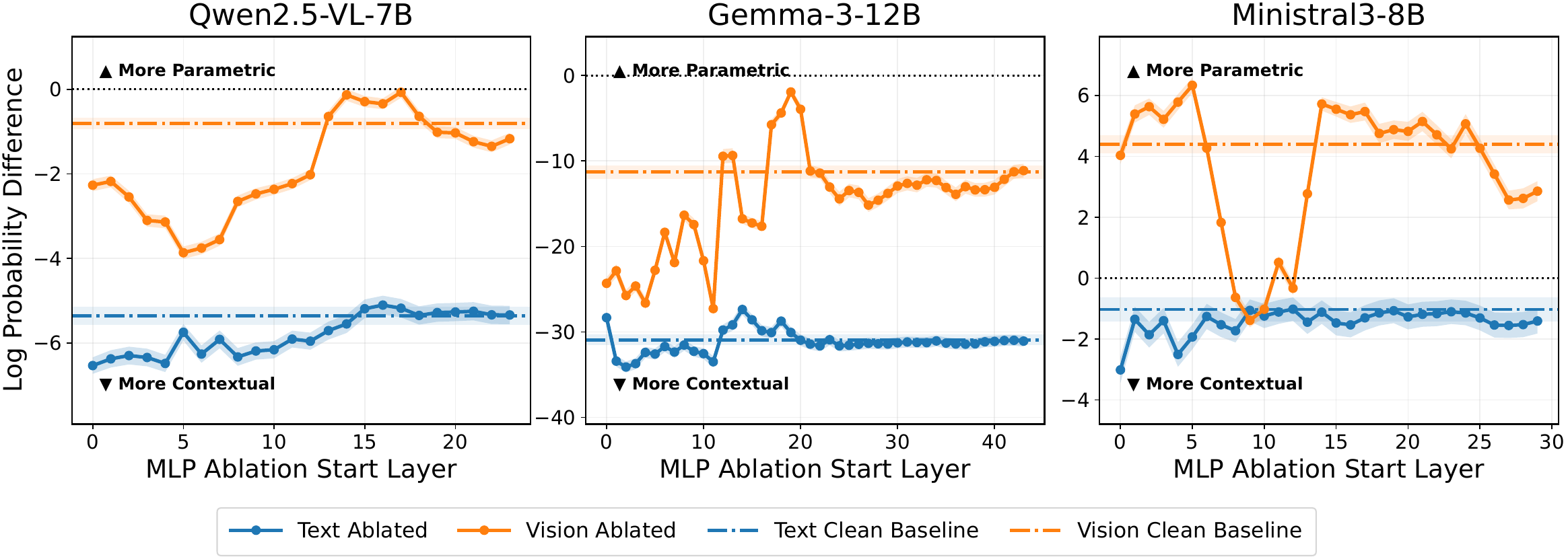}
  \caption{Behavioral impact of MLP ablation across entity positions. The x-axis indicates the starting layer of the ablated MLP span. The y-axis represents the margin defined in Equation \ref{eq:margin}. We utilize a constant span of five layers for all the models. The shaded region represents $\pm$ one standard error of the mean.}
  \label{fig:MLP_Ablate_Results}
\end{figure*}

\begin{figure*}[t]
  \includegraphics[width=\textwidth]{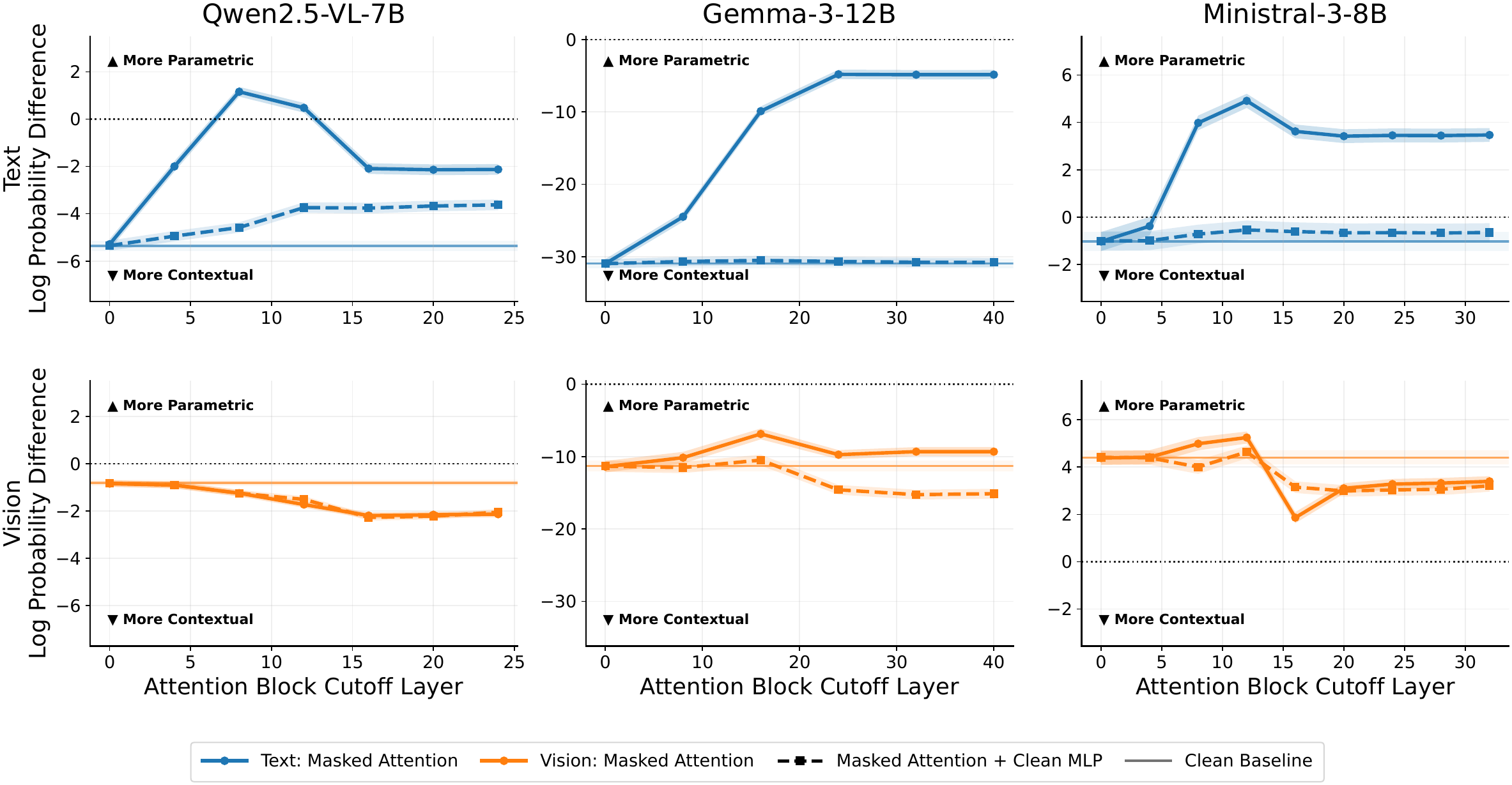}
  \caption{Effect on preference margin when masking attention from entity tokens to the retrieved context. For each cutoff layer $k$ on the x-axis, multi-head self-attention modules from layers $0$ through $k$ were masked so that entity-token positions could not attend to the retrieved context. Solid lines show the standard masking intervention, where downstream entity-token MLPs receive the attention-masked residual stream as input. Dashed lines show the control intervention, where the entity-token MLP outputs added to the residual stream are replaced with their corresponding outputs from the clean, unmasked run. Dash-dotted horizontal lines denote the clean-run margin for each modality. Shaded regions indicate $\pm$ one standard error of the mean (SEM).}
  \label{fig:Attn_Block_Results}
\end{figure*}

We find that ablating entity token MLPs narrows the modality gap by forcing the model to rely on context for visual entities, while leaving textual entities relatively unaffected. As past work on factual recall has identified, the task largely relies on the additive effect of multiple Multilayer Perceptrons (MLPs) within early-to-middle layers over the entity token positions. This process enriches the representation of the subject entity such that the target relation may be extracted by specific “retrieval heads” at the generation token (\citealp{meng2022locating}; \citealp{chughtai2024summing}; \citealp{geva2023dissecting}). Drawing inspiration from this line of work, we adopt a similar causal intervention approach to evaluate whether early MLPs drive this parametric preference under conflict. Specifically, we apply a sliding five-layer ablation window over the entity tokens (additional window-size ablations are provided in Appendix~\ref{sec:appendix-windowsize}). Following each ablation, we measure the change in the model's output preference. We quantify this preference using the log-probability margin, $M$, calculated as:

\begin{equation} \label{eq:margin}
    M = \log P(y_{p}) - \log P(y_{c})
\end{equation} 
where $y_{p}$ is the ground-truth parametric answer (generated when no conflicting context is present), and $y_{c}$ is the contextual answer. While this metric may reflect shifts toward unrelated outputs rather than a change in preference, Appendix~\ref{sec:margin_shift} shows that this behavior is minimal in our experiments.

\paragraph{Results}As shown in Figure~\ref{fig:MLP_Ablate_Results}, across all the models and datasets, the ablation of MLPs appears to diminish the parametric bias in visual entities to a significant degree more than the textual entities. For Qwen and Gemma we see that the largest shift towards the contextual output occurs when ablating early MLPs for visual inputs, which implies they contribute heavily towards the parametric output. This is consistent with \citet{toolatetorecall} findings that the early MLPs play a substantial role in the factual recall task, as well as \citet{geva2023dissecting} findings that entity enrichment occurs over MLPs at the entity tokens. For Ministral, we see that the largest shift for visual inputs is concentrated between layers 9-14, which implies a more localized region for enrichment. Yet across all three models, ablating these early-to-middle MLPs reduces the visual parametric bias toward the textual baseline. This pattern suggests that MLPs for visual entities drive parametric preference, but play a significantly less important role for textual entities.

\subsection{Attention Suppresses MLPs When Sources Conflict}
\label{sec:mechanism:attn}

While MLPs appear responsible for retrieving information about image entities, they have less of an effect on text entities. However, in standard (no-conflict) factual recall, Entity MLPs promote parametric information for both visual and textual entities \citep{toolatetorecall}. To account for this divergence, we consider two potential explanations: 

E1. Baseline asymmetry. MLP-based parametric promotion is inherently weaker for textual entities than for visual ones. Whenever in-context information is available, attention heads can overwrite the entity tokens with contextual information, overriding the (weak) default parametric enrichment. 

E2. Context-sensitive suppression. Parametric promotion is comparably strong across modalities at baseline, but for textual entities, early-layer attention to the context actively suppresses the downstream Entity MLPs — effectively switching off parametric promotion — which is why in-context information prevails. 

\paragraph{Experiment} To investigate how attention over the entity tokens affects the model output, we devise an attention masking experiment where we selectively block contextual information flow into the entity representation. For each transformer layer $l \in \{0, \dots, k\}$, we modify the attention computation such that entity-token queries cannot attend to retrieved context tokens, while all other attention pathways remain unchanged.

To determine whether this behavioral shift is driven directly by the masked attention heads or indirectly through their downstream effects on MLPs, we follow the causal mediation approach of \citet{NEURIPS2020_92650b2e}. We first apply the same attention mask while patching the residual-stream contributions of the entity-position MLPs back to their clean values from an unmodified forward pass. This intervention preserves the direct effect of the attention mask while blocking its downstream effect on MLP-mediated parametric enrichment. We then compare this run against the fully masked intervention, in which downstream MLP activations are allowed to change. The difference between these conditions isolates the extent to which the shift in model preference is mediated by downstream MLPs, rather than by the attention heads alone.

\begin{figure*}[t]
  \includegraphics[width=\textwidth]{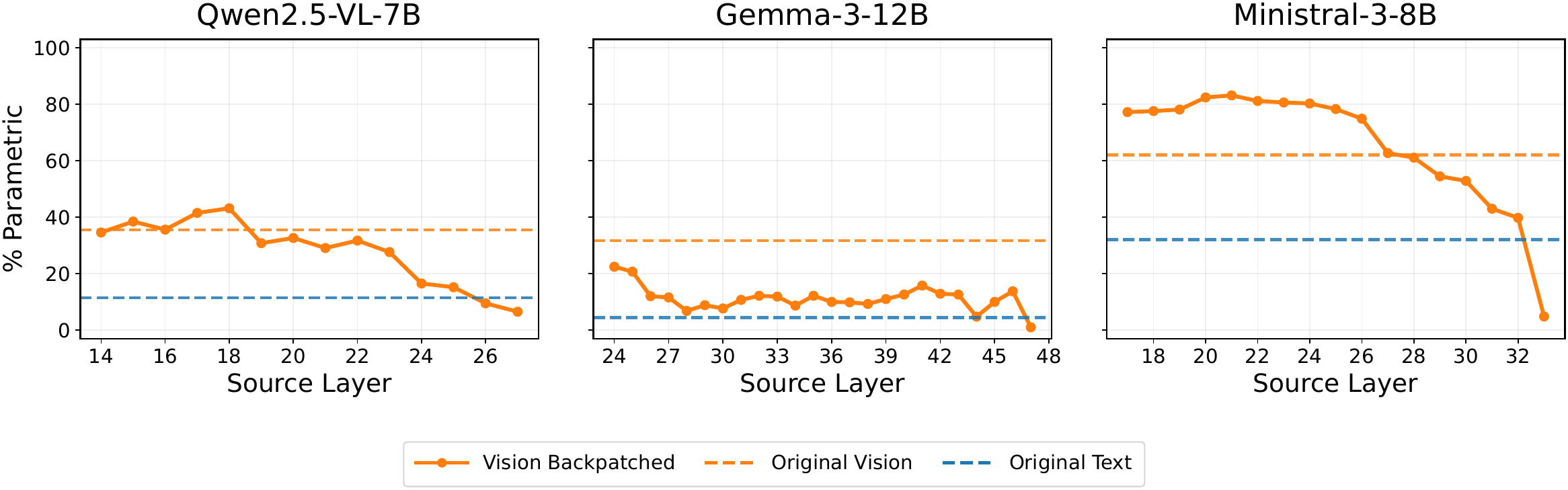}
  \caption{Parametric answer reporting rate under back-patching. We patch activations from the visual-token span of a specific source layer (ranging from layer $\frac{L}{2}$ to the final layer $L$, shown on the x-axis) into the first layer. The solid orange line tracks the model’s resulting parametric answer reporting rate (\%). For comparison, the dashed orange and dashed blue horizontal lines represent the baseline visual and textual parametric reporting rates, respectively.}
  \label{fig:backpatch}
\end{figure*}

\paragraph{Results} The results strongly support E2 (context-sensitive suppression) over E1 (baseline asymmetry). Figure~\ref{fig:Attn_Block_Results} shows that masking attention to the retrieved context over the entity tokens results in a large parametric shift for textual entities. Notably, this shift is heavily mediated by downstream MLPs across all models, as restoring their clean activations largely neutralizes the parametric spike.

In contrast to text, masking attention over visual entities produces only a small shift, suggesting that visual representations are less affected by the conflicting context. Moreover, restoring the clean MLP outputs produces minimal additional change beyond the attention-masking intervention itself. Taken together, this suggests that the context does not strongly suppress parametric enrichment for visual entities.

Interestingly, in many cases, the parametric shifts peak around the middle layers. Masking the later layers causes the preference to drop slightly from its highest point, but it still stays well above the baseline. This suggests that late-layer attention to the supplied context may occasionally support the parametric answer. We hypothesize that this could be due to an entity-enrichment process that occurs in the context, such that late attention to the context can promote parametric information. Overall, however, the main takeaway remains that textual entities exhibit a much stronger dependence on context-attention for suppressing downstream Entity MLP contributions, whereas visual entities are comparatively robust to this suppression.

\subsection{Slow Image Resolution Prevents MLPs From Being Suppressed}
\label{sec:mechanism:proc-time}
 
We hypothesize that failure of MLP suppression for visual entities is due to processing time; specifically, visual features may require more layer depth to resolve into text-aligned concepts, allowing them to escape early-layer suppression. To test this late entity alignment hypothesis, we utilize back-patching, an intervention used in past work to provide additional processing depth for late-resolved representations in LLMs (\citealp{biran-etal-2024-hopping}; \citealp{lepori-etal-2025-racing}) and to re-align mismatched visual and textual processing streams in VLMs (\citealp{toolatetorecall}; \citealp{sametaskdiffcircuit}). This method is motivated by the finding that visual representations increasingly align with text concepts as they progress through later layers (\citealp{venhoff2025visualrepresentationsmaplanguage}; \citealp{wu2025semantichubhypothesislanguage}; \citealp{masry2025alignvlmbridgingvisionlanguage}). 

We test whether patching activations across image tokens from a source layer $S$ into the first layer reduces parametric outputs. If the features at layer $S$ have successfully resolved into text-aligned concepts, back-patching should allow the visual tokens to attend to the conflict and suppress the parametric signal. Furthermore, we extract these clean activations from a standard run with no conflicting context to ensure that the visual entity representations are uncorrupted. Following back-patching, we measure the change in the percentage of parametric outputs to track this behavioral shift.

\paragraph{Results} As shown in Figure~\ref{fig:backpatch}, back-patching the visual entity representations into the first layer drives a strong shift away from the parametric answer and toward the text-consistent contextual response. Across all models, back-patching from later layers decreases the visual parametric preference, ultimately dropping near the text baseline. For Ministral, the effect is less pronounced in the middle layers, though the gap begins to narrow in the later layers. In contrast, Gemma reaches the textual baseline much earlier, showing little further decrease when patching from subsequent layers. These findings suggest that the behavioral divergence under conflict stems from a lack of early alignment between the representations of the two modalities. 

\section{Prompt-Based Methods For Mitigating the Modality Gap}
\label{sec:mitigation}

\begin{figure*}[t]
  \includegraphics[width=\textwidth]{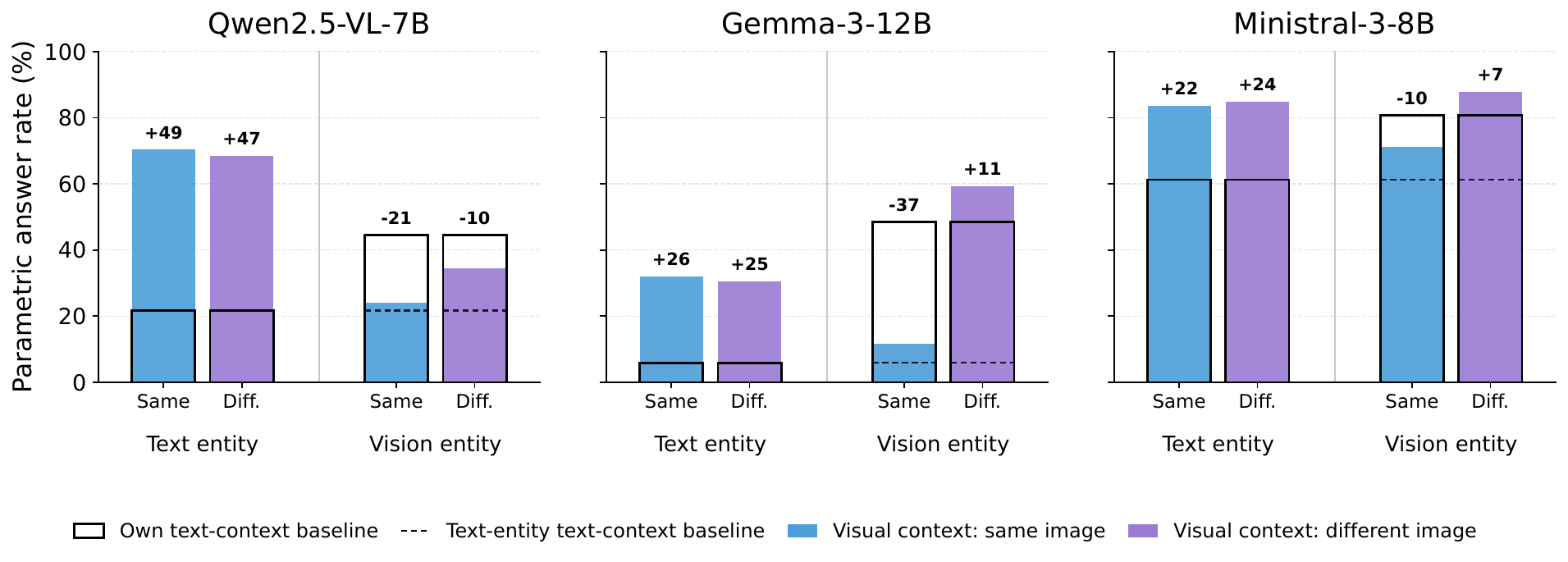}
  \caption{Effect of visual-context prompting on parametric answer rates under context-memory conflict.
Results are aggregated across the building-location and celebrity-career datasets. For each model, bars are grouped by queried entity modality. The black outline shows the standard textual-context baseline for that modality, while the filled bar shows the visual-context result using either the same image or a different image of the entity. Numbers above bars report the change in parametric rate relative to the corresponding baseline. For vision-entity bars, the dashed horizontal marker shows the textual-entity textual-context baseline.}
  \label{fig:vis_ctxt}
\end{figure*}

While back-patching demonstrates that the modality gap can be mitigated through internal activation alignment, we investigate whether prompt-based, ``black-box'' interventions can induce similar convergence across modalities. Specifically, we explore two prompting strategies: Chain-of-Thought (CoT) prompting and visual context simulation.

\subsection{Chain-of-Thought Does Not Bridge the Modality Gap}

\begin{table}[t]
\centering
\footnotesize
\renewcommand{\arraystretch}{0.9}

\resizebox{\columnwidth}{!}{%
\begin{tabular}{lccc}
\toprule
\textbf{Model}
& \textbf{Text CoT (\%)}
& \textbf{Vision CoT (\%)}
& $\Delta$ \\
\midrule

Qwen2.5-VL-7B
& 15.34 (+3.91)
& 42.64 (+7.08)
& 27.30 {\color{red}$\uparrow$ +3.17} \\

Gemma-3-12B
& 15.17 (+10.70)
& 31.12 (-0.56)
& 15.94 {\color{ForestGreen}$\downarrow$ -11.27} \\

Ministral-3-8B
& 2.50 (-29.55)
& 33.14 (-28.88)
& 30.63 {\color{red}$\uparrow$ +0.66} \\

\bottomrule
\end{tabular}%
}

\caption{
Effect of Chain-of-Thought (CoT) prompting under context-memory conflict. Values denote the percentage of parametric answers.
Parenthetical values indicate the change relative to the Baseline runs.
$\Delta$ denotes the modality gap (Vision -- Text).
}
\label{tab:cot_results}
\end{table}
 Past work investigating apparent modality gaps in VLMs has utilized Chain-of-Thought (CoT) prompting to mitigate divergent behavior (\citealp{toolatetorecall}; \citealp{cotinvis}). CoT may help VLMs resolve visual entities before reasoning through the prompt more similarly to text. However, as shown in Table~\ref{tab:cot_results}, evaluating the models with CoT prompting (see Appendix~\ref{sec:prompting_details} for prompt template) yields relatively inconsistent results across models. In Qwen, we see that the CoT prompting has little effect on the model's output, whereas in Ministral, both visual and textual entities display a noticeable decrease in parametric responses. Despite this drop, the divergence between modalities remains similar. Gemma did see the gap decrease slightly, as CoT induced more parametric reporting in text cases. Overall, we observe that a similar behavioral divergence persists across all three models. While these results indicate that CoT cannot be reliably leveraged to enforce task consistency, they further support our findings in Section~\ref{sec:mechanism:attn}: parametric bias may stem from a lack of parametric suppression, rather than a comparatively weakened contextual binding.
 
\subsection{Visual Context Can Prevent Textual Suppression}
\label{sec:vctxt_med_gap}

Our previous experiments have shown that textual entities display a contextual bias due to early resolution and contextual integration; we attempt to alter this process through prompting. Specifically, we replace the context textual reference with an image. For example, instead of stating ``The Eiffel Tower is located in Tokyo'', we provide an image of the Eiffel Tower and state ``The entity pictured is located in Tokyo''. By requiring the context to first resolve the image, we hypothesize that many of the early entity-enriching MLPs in the textual setting will no longer be suppressed, switching those answers back to parametric. Conversely, when the image in the query matches the image in the context, we expect the model to easily connect them. This clear representation alignment should allow the visual entity to detect the conflict and suppress the MLPs, switching the final answer from parametric to contextual. To evaluate the robustness of this effect, we test two conditions: one using the identical image from the visual query, and another using a different image of the same entity. Furthermore, we limit this evaluation to the celebrity and building datasets, as sourcing multiple images for unique historical artworks is fundamentally constrained.

As shown in Figure~\ref{fig:vis_ctxt}, replacing the textual context with an image increases the parametric response rate for text entities across all models, supporting our hypothesis. When we query visual entities using the exact same image found in the context, the parametric rate drops for all three models compared to the baseline. Furthermore, the parametric rate approaches the baseline rate observed for text entities, suggesting that querying with a representation closely aligned to the context increases the model’s likelihood of accepting the contextual information. In Appendix~\ref{sec:vis_ctxt_mech}, we causally validate that MLP suppression mediates this inverse behavior. However, this effect weakens when we switch to a different image of the same entity. These results further support our mechanistic findings that context-driven suppression requires an early aligned representation between the queried entity and the contextual information. This inconsistency implies that the suppression mechanism relies on precise representational alignment, making it fragile to visual variations of the same entity.

\section{Conclusion}

Our research investigates whether context-memory conflicts are resolved consistently across the modalities of a prompted entity. We find that across various models and datasets, Vision-Language Models rely on parametric knowledge at a noticeably higher rate when prompted with visual entities compared to their textual counterparts. 

Our causal interventions reveal a clear mechanistic divide. For textual entities, early-layer attention to conflicting context actively suppresses parametric promotion within downstream MLPs, allowing the contextual information to prevail. In contrast, the parametric signal from visual entity tokens remains relatively invariant to this suppression. By leveraging back-patching to shift fully resolved, late-layer visual features into the initial layers, we induce behavior more consistent with textual counterparts. Coupled with our prompt-based approaches, these results suggest that the modality divergence is rooted in a failure to suppress parametric retrieval when the in-context and queried entity representations do not align in early layers.

These findings provide a critical first step toward ensuring consistent conflict resolution in VLMs. We hope this work inspires future grounding methods and Retrieval-Augmented Generation (RAG) frameworks to account for these inherent modal biases when integrating external knowledge into multimodal systems.

\section*{Limitations}
Our study investigates context-memory conflicts in a controlled mock-rag setting to limit confounding variables in our mechanistic analysis. Consequently, the experimental setup does not capture the diversity of real-world usage, as models may prepend visual entities, and users may interleave multimodal information across complex, multi-image document layouts. Furthermore, our study's scope is focused on open-source vision-language models. While we observe consistent behavioral trends across model families, it remains unclear whether the same mechanisms scale to substantially larger proprietary systems. Future research should expand on this direction by investigating model behavior within true text and multimodal RAG systems to determine how these mechanistic differences in conflict resolution affect the reliability of VLMs in real-world deployments.

\section*{Acknowledgments}

We thank the members of the Language Understanding and Representation
(LUNAR) Lab at Brown University – especially Michael A.
Lepori, Zhuonan Yang, and other Brown SOLAR members for their helpful discussion and
feedback on our work. This project was supported in part by the Young Faculty Award from the Defense Advanced Research Projects Agency Grant \#D24AP00261, the Schmidt Sciences Grant \#GR5300958, and the NSF AI Research Institute on Interaction for AI Assistants Grant \#GR5300593. Ellie Pavlick is a paid consultant for Google DeepMind. The content of this article does not necessarily reflect the views of the US Government or of Google, and no official endorsement of this work should be inferred.

\section*{Ethical Considerations}
Our study investigates how and why vision-language models resolve context-memory conflicts differently depending on the input modality. In our investigation, we provide showcase interventions and behavioral methods that make these models more receptive to in-context information. While this can ensure AI systems are up to date in their knowledge when paired with a trustworthy RAG, it also induces the risk of these systems outputting false claims or information. Additionally, our experiments use publicly available entities and factual attributes, and do not introduce new private or sensitive personal information. However, because some examples involve real people and landmarks, future work should consider potential harms from incorrect attribution, outdated facts, or misleading contextual claims. 

\newpage

\bibliography{custom}
\clearpage
\appendix

\section{Dataset Curation}
\label{sec:appendix-dataset}

To study how vision-language models resolve context-memory conflicts at scale we curate a multimodal conflict dataset inspired by the format of the MMKC-Bench dataset \citep{Jia_Du_Jiang_Liang_Ren_Xin_Yang_Feng_Chen_Lu_etal._2026}. We curated our dataset from three domains: celebrities, buildings, and artworks (see Table~\ref{tab:benchmark}). 

\subsection{Domain Curation Pipelines}

\paragraph{Celebrity}
We expanded the MMKC-Bench celebrity context-memory conflict dataset from 149 celebrities to 476 celebrities. We discovered additional public figure candidates by querying English Wikipedia page views to target high-confidence individuals. For each candidate, we pulled available metadata from Wikidata, extracting their date of birth, citizenship, and primary occupation. To balance the dataset, we applied a ranking preference which filtered out generic performance occupations (such as actors or singers) in order to have a more mixed set of professions such as scientists, politicians, and professional athletes. High-resolution ground-truth images were then downloaded from the main Wikipedia infoboxes.

\paragraph{Artwork}
We used a lightweight Wikidata SPARQL query to discover distinct artwork categorized specifically as paintings or sculptures. For each artwork, we extracted three attributes: the creator, the creation year, and the country where the piece is currently located. Conflicting counterparts for this subset were drawn from a randomized set of 20 artists and 20 countries. Artwork was sourced directly from Wikidata image claims and Wikimedia Commons file paths. 

\paragraph{Building}
We started with the clean training subset of the Google Landmarks v2 (GLDv2) dataset. To isolate human-made landmarks, we filtered categories using hierarchical keyword matching (e.g., keeping titles containing \textit{tower, cathedral, bridge, or palace} while strictly dropping natural features like \textit{lakes or mountains}). We mapped the remaining Wikimedia category URLs to explicit Wikidata items, automatically following main topic links to resolve core entities. Finally, we extracted the corresponding architect, completion year, and country location for each landmark. Conflicts were generated by randomly sampling from a pool of 20 countries, a $\pm$ 20-year range, and 20 renowned architects. 

\subsection{Conflict Format and MCQ Generation}

\begin{figure}[t]
\centering

\begin{tcolorbox}[
    colback=white,
    colframe=black,
    boxrule=0.5pt,
    arc=1pt,
    left=5pt,
    right=5pt,
    top=5pt,
    bottom=5pt,
    width=0.98\linewidth
]

\scriptsize
\begin{verbatim}
<bos><start_of_turn>user

Context information is below.
---------------------
The person pictured was born in 1989.
The person pictured is best known as a novelist.
---------------------
Given the context information and your knowledge,
answer the question about the following entity.
Entity:

<start_of_image>
<image_soft_token>
<end_of_image>

Query: What is this person most famous for?
Answer with only 1-3 word(s).
<end_of_turn>

<start_of_turn>model
Answer:
\end{verbatim}

\end{tcolorbox}

\caption{Representative raw prompt template for the visual-context setting, shown with the original chat formatting and special image tokens used by the model family.}
\label{fig:raw_prompt_example}

\end{figure}

For every entity in our dataset, we compiled a standardized evaluation block containing three components:
\begin{enumerate}
    \item \textbf{Ground-Truth Knowledge:} A factual paragraph combining the three retrieved metadata attributes into a clean string (e.g., \textit{``[Entity] was designed by [Architect] in [Year].''}).
    \item \textbf{Counterfactual Mis-knowledge:} A set of modified text paragraphs where a single target attribute is replaced with a misleading distractor.
    \item \textbf{Evaluation Queries:} Paired open-ended queries and deterministic 4-choice multiple-choice questions (MCQs) designed to test the model's reliance on text context versus internal memory. While we curated MCQ queries, all of our experiments throughout this paper were conducted using the open-ended queries.
\end{enumerate}
To generate high-quality multiple-choice options, alternative distractors were sampled from a global pool of valid in-domain attributes (e.g., pulling alternative historical years or architects from other rows in the dataset). This ensured that all wrong options remained highly plausible. The simulated supplied context presented facts in a consistent order as described below.
\begin{itemize}
    \item \textbf{Celebrity: [Entity] was born on [Birthday]. [Entity] is [Nationality]. [Entity] is a [Occupation].} 
    \item \textbf{Building:} [Entity] was designed by [Architect]. [Entity] was completed in [Year]. [Entity] is located in [Country].

    \item \textbf{Artwork:} [Entity] was created by [Artist] in [Year]. [Entity] is currently located in [Country].
\end{itemize}
Additionally, the queried entity is formatted cleanly within the prompt based on the target domain. For celebrity queries, the text simply lists the individual's full name. For the other domains, we wrap the identifier with structural context to ensure clarity for the models: building entities are prefixed as \textit{``The building [EntityName]''}, artworks are formatted as \textit{``The artwork titled: [ArtworkName AND Year Completed]''}, and logos are stated as ``The logo of the company known as [CompanyName]''. For artwork we included the completion year in the textual case as we observed that factual recall performance suffered compared to the visual case without it. An example of the raw template can be seen in Figure~\ref{fig:raw_prompt_example}.

\subsection{Model Specific Dataset Filtering}
\label{sec:Dataset_filtering}

To guarantee that our context-memory conflict experiments ensured the model had to address a conflict between a memorized fact and the context, rather than a baseline failure to recognize an entity or recall a fact, we apply a strict two-stage data filter for each model (results in Table~\ref{tab:filter_pipeline}).

First, we verify that the model can identify the entity in the image. We query the model using a standard prompt tailored to the domain: \textit{``Name the person in this image.''}, \textit{``Name the building in this image.''}, or \textit{``Name the artwork in this image.''}. We then apply regular expressions to normalize whitespaces, strip casing, and remove punctuation from the model's text response. A candidate entry is kept only if the normalized response yields a successful string match with the target entity's ground-truth name.

Second, for all entities that pass visual recognition, we evaluate whether the model successfully possesses the relevant internal parametric memory when prompted with both the textual and visual entities. We query the model across both modalities with open-ended factual questions corresponding to the specific relation under study (e.g., querying for a building's location). To evaluate accuracy while accounting for variations in formatting, paraphrasing, or phrasing, we use an LLM-as-a-judge setup leveraging \texttt{Meta-Llama-3-8B-Instruct} \cite{llama3modelcard}. The judge model is prompted to score the response strictly as \texttt{CORRECT} or \texttt{INCORRECT} against the ground-truth attribute, treating refusals (\textit{``I don't know''}) as incorrect. Only entities that satisfy both visual recognition and correct factual recall are retained for our final evaluation datasets.

We observe that the building subcategories other than location contain significantly fewer queries after filtering. Nevertheless, we report results for all subcategories rather than aggregating over all the subcategories. Doing so is important because the retention rate itself provides insight into the strength of the model’s underlying parametric knowledge. Relations with the largest retained datasets correspond to attributes that are core relations of the entities (e.g., celebrity careers, building locations, and artwork artists). Notably, these same relations also exhibit the strongest modality divergence in our context-memory conflict evaluations. Reporting all subcategories therefore allows us to distinguish between weakly represented attributes, where models often fail baseline factual recall altogether, and core entity attributes, where models possess strong internal knowledge yet still resolve contextual conflicts differently across modalities.

\begin{table*}[t]
\centering
\small
\setlength{\tabcolsep}{5pt}
\renewcommand{\arraystretch}{0.92}
\begin{tabular}{llrrrr}
\toprule
\textbf{Model} & \textbf{Relation} & \textbf{Potential} & \textbf{After Entity} & \textbf{After Fact Recall} & \textbf{Retained (\%)} \\
\midrule
\multirow{8}{*}{\rotatebox{90}{Qwen2.5-VL-7B}}
& \textbf{Celebrity -- Occupation} & \textbf{952} & \textbf{418} & \textbf{328} & \textbf{34.45} \\
& Celebrity -- Birth year & 952 & 418 & 142 & 14.92 \\
& Celebrity -- Nationality & 952 & 418 & 308 & 32.35 \\
& Building -- Completion year & 5,410 & 522 & 90 & 01.66 \\
& Building -- Architect & 5,410 & 522 & 24 & 00.44 \\
& \textbf{Building -- Location} & \textbf{5,410} & \textbf{522} & \textbf{454} & \textbf{08.39} \\
& \textbf{Artwork -- Artist} & \textbf{9,442} & \textbf{1,020} & \textbf{228} & \textbf{02.41} \\
& Artwork -- Location & 9,442 & 1,020 & 368 & 03.90 \\
\cmidrule(lr){1-6}
\textbf{Total} & -- & \textbf{37,970} & \textbf{4,860} & \textbf{1,942} & \textbf{05.11} \\
\midrule
\multirow{8}{*}{\rotatebox{90}{Qwen2.5-VL-32B}}
& \textbf{Celebrity -- Occupation} & \textbf{952} & \textbf{394} & \textbf{312} & \textbf{32.77} \\
& Celebrity -- Birth year & 952 & 394 & 152 & 15.97 \\
& Celebrity -- Nationality & 952 & 394 & 260 & 27.31 \\
& Building -- Completion year & 5,410 & 692 & 132 & 02.44 \\
& Building -- Architect & 5,410 & 692 & 82 & 01.52 \\
& \textbf{Building -- Location} & \textbf{5,410} & \textbf{692} & \textbf{638} & \textbf{11.79} \\
& \textbf{Artwork -- Artist} & \textbf{9,442} & \textbf{878} & \textbf{330} & \textbf{03.50} \\
& Artwork -- Location & 9,442 & 878 & 216 & 02.29 \\
\cmidrule(lr){1-6}
\textbf{Total} & -- & \textbf{37,970} & \textbf{5,014} & \textbf{2,122} & \textbf{05.59} \\
\midrule
\multirow{8}{*}{\rotatebox{90}{Qwen2.5-VL-72B}}
& \textbf{Celebrity -- Occupation} & \textbf{952} & \textbf{436} & \textbf{334} & \textbf{35.08} \\
& Celebrity -- Birth year & 952 & 436 & 244 & 25.63 \\
& Celebrity -- Nationality & 952 & 436 & 330 & 34.66 \\
& Building -- Completion year & 5,410 & 532 & 152 & 02.81 \\
& Building -- Architect & 5,410 & 532 & 136 & 02.51 \\
& \textbf{Building -- Location} & \textbf{5,410} & \textbf{532} & \textbf{472} & \textbf{08.72} \\
& \textbf{Artwork -- Artist} & \textbf{9,442} & \textbf{1,140} & \textbf{544} & \textbf{05.76} \\
& Artwork -- Location & 9,442 & 1,140 & 388 & 04.11 \\
\cmidrule(lr){1-6}
\textbf{Total} & -- & \textbf{37,970} & \textbf{5,184} & \textbf{2,600} & \textbf{06.85} \\
\midrule
\multirow{8}{*}{\rotatebox{90}{Gemma-3-12B}}
& \textbf{Celebrity -- Occupation} & \textbf{952} & \textbf{322} & \textbf{228} & \textbf{23.95} \\
& Celebrity -- Birth year & 952 & 322 & 204 & 21.43 \\
& Celebrity -- Nationality & 952 & 322 & 252 & 26.47 \\
& Building -- Completion year & 5,410 & 256 & 64 & 01.18 \\
& Building -- Architect & 5,410 & 256 & 32 & 00.59 \\
& \textbf{Building -- Location} & \textbf{5,410} & \textbf{256} & \textbf{224} & \textbf{04.14} \\
& \textbf{Artwork -- Artist} & \textbf{9,442} & \textbf{852} & \textbf{270} & \textbf{02.86} \\
& Artwork -- Location & 9,442 & 852 & 254 & 02.69 \\
\cmidrule(lr){1-6}
\textbf{Total} & -- & \textbf{37,970} & \textbf{3,438} & \textbf{1,528} & \textbf{04.02} \\
\midrule
\multirow{8}{*}{\rotatebox{90}{Gemma-3-27B}}
& \textbf{Celebrity -- Occupation} & \textbf{952} & \textbf{358} & \textbf{242} & \textbf{25.42} \\
& Celebrity -- Birth year & 952 & 358 & 268 & 28.15 \\
& Celebrity -- Nationality & 952 & 358 & 244 & 25.63 \\
& Building -- Completion year & 5,410 & 326 & 130 & 02.40 \\
& Building -- Architect & 5,410 & 326 & 86 & 01.59 \\
& \textbf{Building -- Location} & \textbf{5,410} & \textbf{326} & \textbf{306} & \textbf{05.66} \\
& \textbf{Artwork -- Artist} & \textbf{9,442} & \textbf{1,042} & \textbf{420} & \textbf{04.45} \\
& Artwork -- Location & 9,442 & 1,042 & 354 & 03.75 \\
\cmidrule(lr){1-6}
\textbf{Total} & -- & \textbf{37,970} & \textbf{4,136} & \textbf{2,050} & \textbf{05.40} \\
\midrule
\multirow{8}{*}{\rotatebox{90}{Ministral-3-8B}}
& \textbf{Celebrity -- Occupation} & \textbf{952} & \textbf{268} & \textbf{184} & \textbf{19.33} \\
& Celebrity -- Birth year & 952 & 268 & 34 & 03.57 \\
& Celebrity -- Nationality & 952 & 268 & 162 & 17.02 \\
& Building -- Completion year & 5,410 & 194 & 32 & 00.59 \\
& Building -- Architect & 5,410 & 194 & 10 & 00.18 \\
& \textbf{Building -- Location} & \textbf{5,410} & \textbf{194} & \textbf{178} & \textbf{03.29} \\
& \textbf{Artwork -- Artist} & \textbf{9,442} & \textbf{736} & \textbf{216} & \textbf{02.29} \\
& Artwork -- Location & 9,442 & 736 & 272 & 02.88 \\
\cmidrule(lr){1-6}
\textbf{Total} & -- & \textbf{37,970} & \textbf{2,858} & \textbf{1,088} & \textbf{02.87} \\
\midrule
\multirow{8}{*}{\rotatebox{90}{Ministral-3-14B}}
& \textbf{Celebrity -- Occupation} & \textbf{952} & \textbf{284} & \textbf{204} & \textbf{21.43} \\
& Celebrity -- Birth year & 952 & 284 & 100 & 10.50 \\
& Celebrity -- Nationality & 952 & 284 & 198 & 20.80 \\
& Building -- Completion year & 5,410 & 206 & 46 & 00.85 \\
& Building -- Architect & 5,410 & 206 & 38 & 00.70 \\
& \textbf{Building -- Location} & \textbf{5,410} & \textbf{206} & \textbf{180} & \textbf{03.33} \\
& \textbf{Artwork -- Artist} & \textbf{9,442} & \textbf{752} & \textbf{262} & \textbf{02.77} \\
& Artwork -- Location & 9,442 & 752 & 266 & 02.82 \\
\cmidrule(lr){1-6}
\textbf{Total} & -- & \textbf{37,970} & \textbf{2,974} & \textbf{1,294} & \textbf{03.41} \\
\bottomrule
\end{tabular}

\caption{
Dataset filtering pipeline used for evaluation. ``Potential'' denotes the total number of context--memory conflict pairs available for each domain and relation category. ``After Entity'' corresponds to remaining pairs after filtering for visual recognition, while ``After Fact Recall'' denotes where the model additionally answered the associated fact-recall question correctly. Retained (\%) is computed relative to the ``Potential'' subset.
}
\label{tab:filter_pipeline}
\end{table*}

\subsection{LLM-as-a-Judge Evaluation}
\label{sec:appendix-judge}

\begin{figure}[t]
\centering

\begin{tcolorbox}[
    colback=white,
    colframe=black,
    boxrule=0.5pt,
    arc=1pt,
    left=5pt,
    right=5pt,
    top=5pt,
    bottom=5pt,
    width=0.98\linewidth
]

\scriptsize
\begin{verbatim}
Task: Classify RESPONSE based on similarity
to two reference answers.

Rules:
1. Output ONLY: PARAMETRIC_ANSWER,
   CONTEXTUAL_ANSWER, or neither.
2. If RESPONSE supports BOTH, output
   CONTEXTUAL_ANSWER.
3. Allow for paraphrasing and synonyms.
4. Refusals ("I don't know") = neither.

Input:
Query: {query}
PARAMETRIC_ANSWER: {parametric_ans}
CONTEXTUAL_ANSWER: {contextual_ans}
RESPONSE: {response}

IS THE RESPONSE MORE SIMILAR TO PARAMETRIC_ANSWER
OR CONTEXTUAL_ANSWER? (RESPOND IN ONE WORD)
\end{verbatim}

\end{tcolorbox}

\caption{LLM-as-a-judge prompt used to classify model outputs under context--memory conflict. The judge assigns each response to the parametric answer, the conflicting contextual answer, or neither, while allowing for paraphrases and synonyms.}
\label{fig:judge_prompt_context_memory}

\end{figure}

Throughout our behavioral experiments, we use an LLM-as-a-judge setup with Meta-Llama-3-8B-Instruct~\cite{llama3modelcard} as our judge. As mentioned in the previous subsection, we use it for filtering our dataset to only include instances where the model could retrieve the ground-truth answer parametrically through factual recall. We also use the judge model to classify responses under context-memory conflict as either supporting the parametric answer, the conflicting contextual answer, or neither.

To account for paraphrasing and formatting variation in open-ended generation, the judge is instructed to classify responses semantically rather than through exact string matching. Refusals (e.g., ``I don't know'') are treated as neither or incorrect depending on the evaluation setting. We decode judge responses deterministically using greedy decoding ($T=0$). Additionally, we normalize capitalization, punctuation, and minor label variations before assigning the final class label.

Our judging prompt for the context-memory conflict setting is displayed in Figure~\ref{fig:judge_prompt_context_memory}.

\section{Effect of Window Size on MLP Ablations}
\label{sec:appendix-windowsize}

\begin{figure*}[t]
  \includegraphics[width=\textwidth]{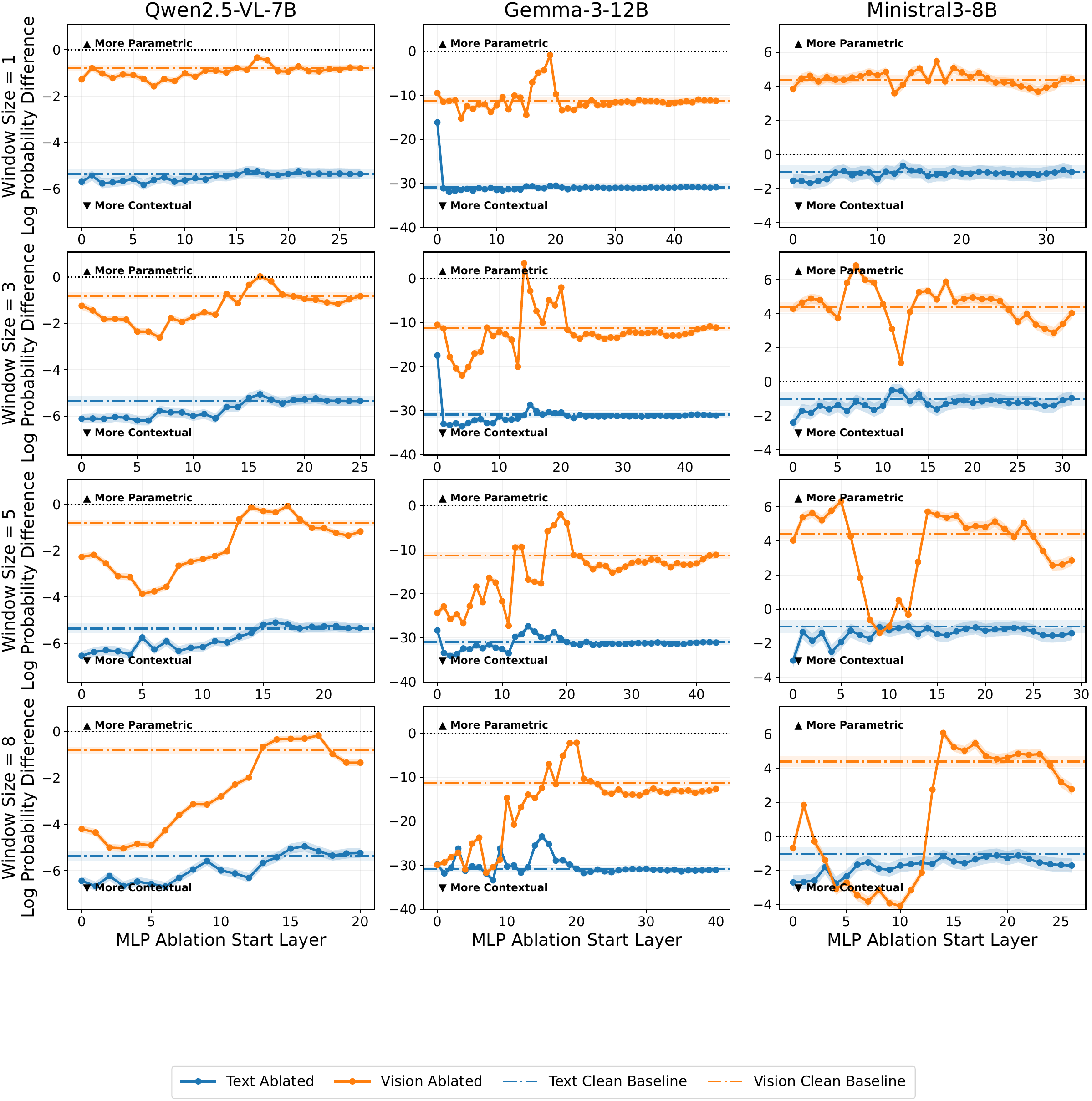}
  \caption{Behavioral impact of MLP ablation across entity positions for different ablation window sizes. The x-axis indicates the starting layer of the ablated MLP span, while each row corresponds to a different window size. The y-axis represents the margin defined in Equation~\ref{eq:margin}. Positive values indicate a preference toward the parametric answer, whereas negative values indicate a preference toward the contextual answer. Dashed horizontal lines denote the clean (non-ablated) baseline behavior. The shaded regions represent $\pm$ one standard error of the mean (SEM).}
  \label{fig:MLP_WS_Results}
\end{figure*}

We test the effect of ablating MLPs over entity positions with window sizes of 1, 3, 5, and 8. Utilizing 2  NVIDIA RTX A5000, our experiments take approximately 24 hours per window size. For visual inputs, the entity tokens are defined by the image patches; for textual inputs, they are defined by the tokens that make up the entity's full name, including contextualizing tokens such as ``The artwork titled...''. Our findings (shown in Figure~\ref{fig:MLP_WS_Results}) show the most apparent shifts for either modality with window sizes of 5 and 8. Interestingly, smaller window sizes show a slight shift towards the parametric answer in the visual cases for Gemma. This suggests that the models may have small windows of MLPs, that promote the contextual information under conflict. Overall, we find that as we increase the window size, we see the trend reported in Section~\ref{sec:mechanism:mlps}: Visual entity MLPs necessitate the final parametric signal, whereas in the textual case, they play a significantly diminished role. 

\section{Observed Margin Shifts Are Not Artifacts of Distribution Shift}
\label{sec:margin_shift}

For the MLP ablation and attention masking experiment in Section~\ref{sec:mechanism}, we utilized a margin metric defined in Equation~\ref{eq:margin}. Specifically, we measure the models' probability of outputting the token sequence associated with both answers. These sequence probabilities were used as a continuous measurement of the parametric and contextual signal. However it is possible that a drop or shift in a direction could be due to the intervention on the model, leading to an out-of-distribution output (I.E. the models output vocabulary space is neither associated with the parametric or contextual answer, but rather is some irrelevant information). 

\subsection{MLP Ablations}

\begin{figure*}[t]
  \includegraphics[width=\textwidth]{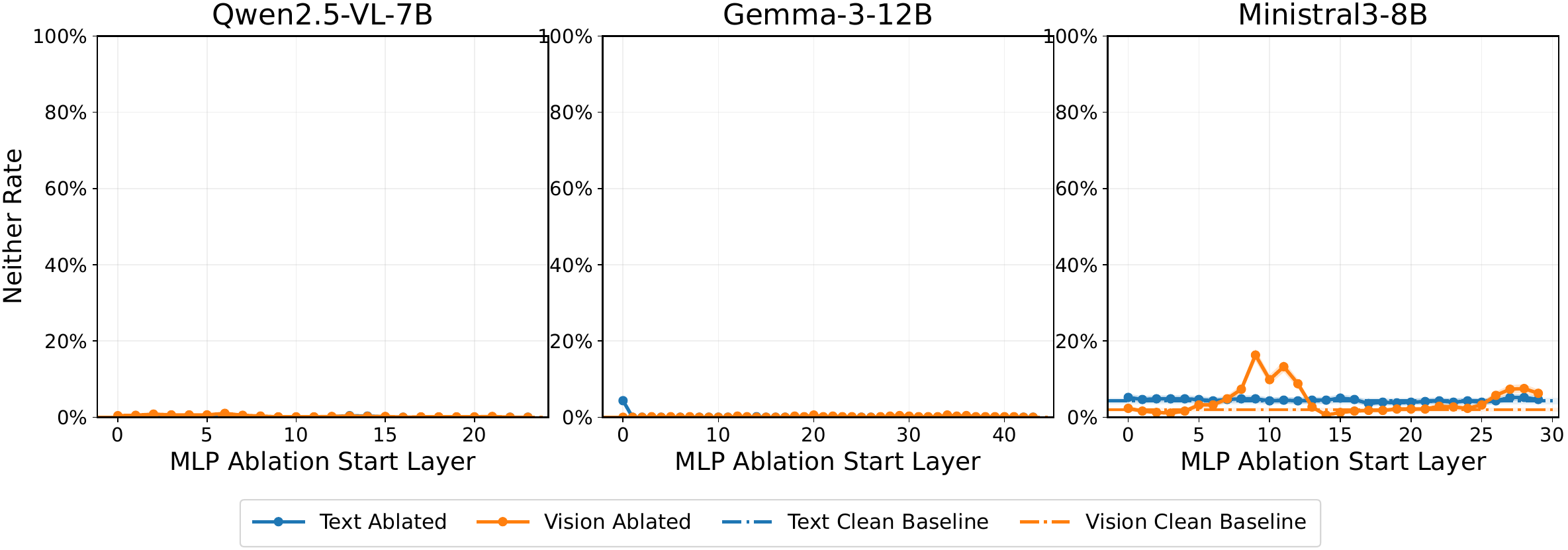}
  \caption{Behavioral impact of MLP ablation on neither-answer generation under context-memory conflict. The x-axis indicates the starting layer of the ablated MLP span, while the y-axis reports the percentage of responses classified as neither the parametric nor contextual answer. Solid curves denote the ablated condition and dash-dotted horizontal lines denote the clean baseline without ablation.}
  \label{fig:MLP_neither_results}
\end{figure*}

\begin{figure*}[t]
  \includegraphics[width=\textwidth]{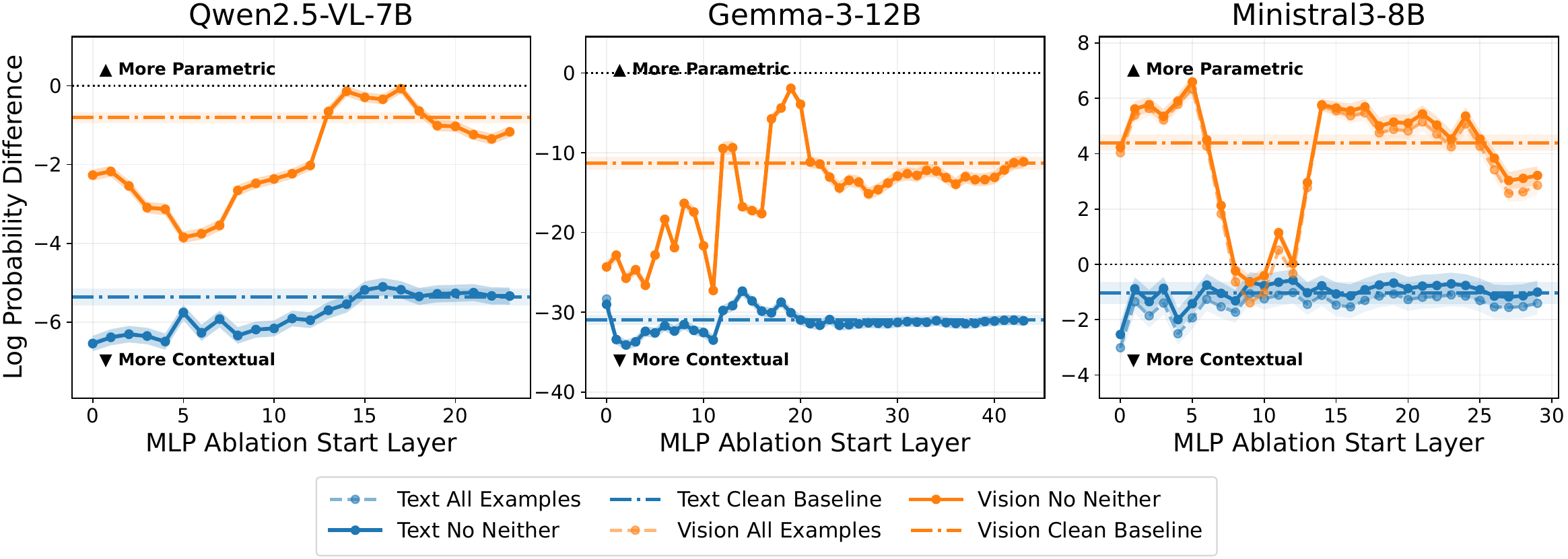}
  \caption{Effect of MLP ablation on the preference margin under context--memory conflict. Solid lines denote the average margin after filtering out responses classified as neither parametric nor contextual, while dashed lines denote the average margin across all responses. Dash-dotted horizontal lines indicate the clean baseline margin without ablation. Positive values correspond to more parametric behavior, while negative values correspond to more contextual behavior.}
  \label{fig:MLP_no_neither_results}
\end{figure*}

We measure the rate at which the MLP ablation interventions results in the model outputting a response that is neither parametric nor contextual (as decided by our LLM-as-a-Judge setup). As shown in Figure~\ref{fig:MLP_neither_results}, the neither rate remains relatively low, suggesting that the interventions primarily shift the model between parametric and contextual behavior rather than inducing unrelated generations. While ministral does show an increase in reporting, neither along the MLPs nor in the parametric shift do we see the same pattern; we confirm that this is not the reason for the shift by validating that the results with ``neither'' cases filtered out show the same trend (Figure~\ref{fig:MLP_no_neither_results}).


\begin{figure*}[t]
  \includegraphics[width=\textwidth]{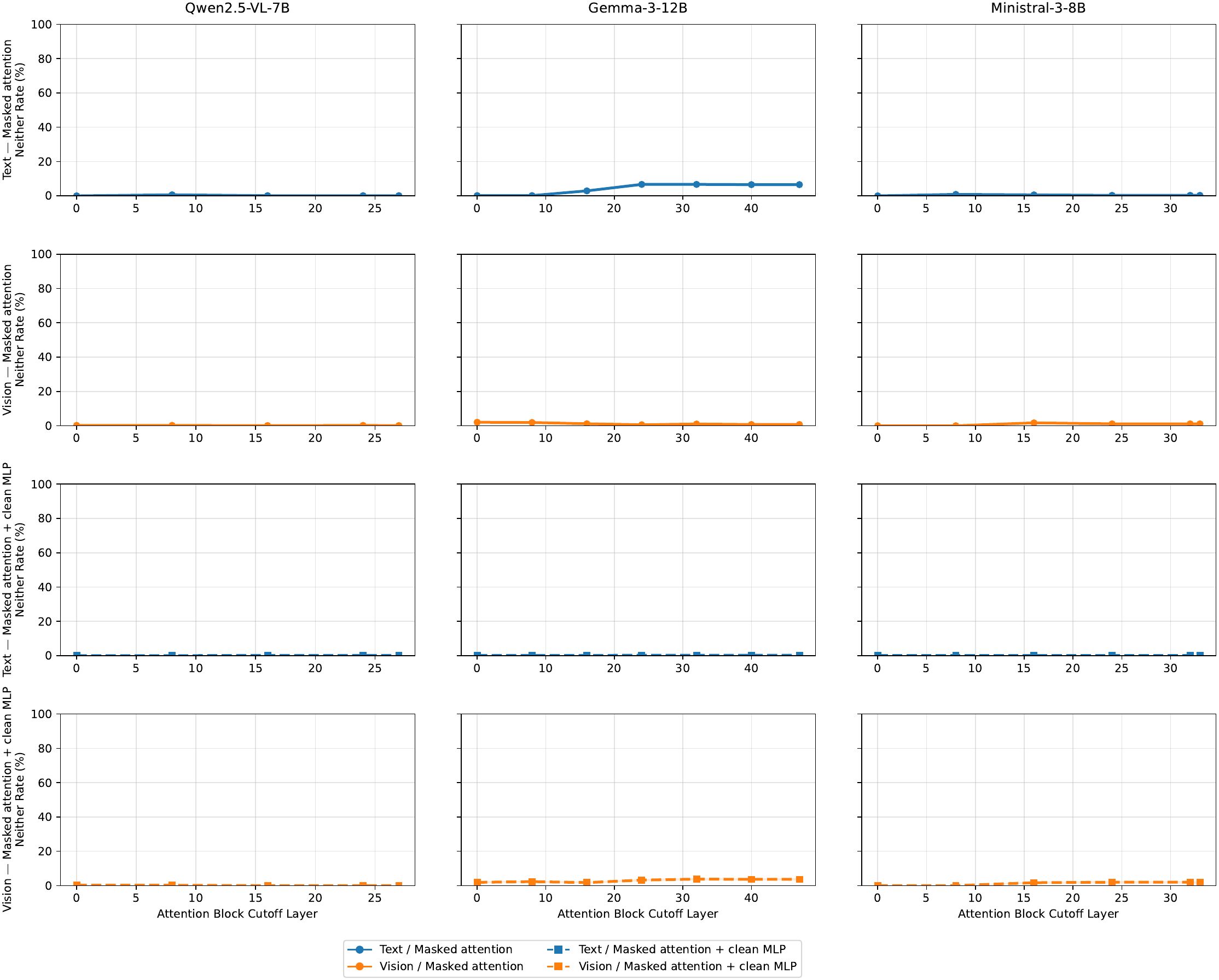}
  \caption{Behavioral impact of the attention-masking experiment. The x-axis indicates the cutoff layer $k$, where attention from entity tokens to retrieved-context tokens is masked for layers $0$ through $k$. The y-axis reports the percentage of responses classified as neither the parametric nor contextual answer. Solid curves denote the masked condition, while dash-dotted horizontal lines indicate the clean, unablated baseline.}
  \label{fig:Attn_mask_neither}
\end{figure*}

\subsection{Attention Masking}

Similarly to the previous subsection, we measure the rate at which the attention masking experiment results in the model outputting a result that is neither parametric nor contextual. As shown in Figure~\ref{fig:Attn_mask_neither}, neither rate stays below 10\% across all models, implying that the interventions do not lead to out-of-distribution outputs.

\begin{figure*}[t]
  \includegraphics[width=\textwidth]{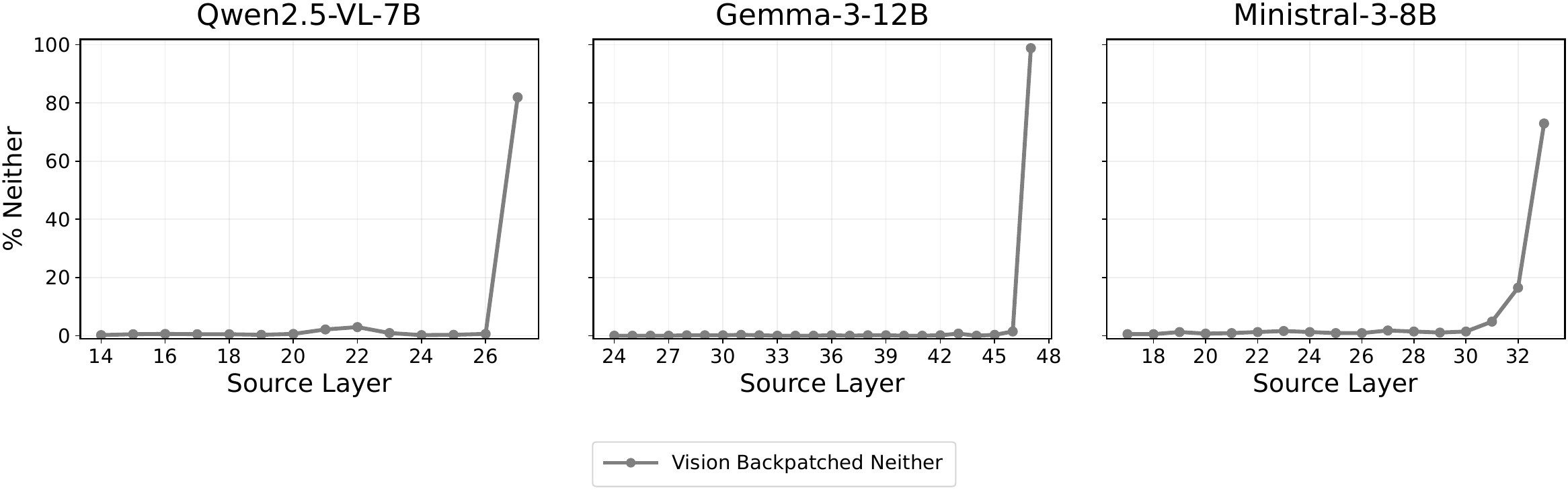}
  \caption{Neither reporting rate under back-patching. After back-patching from source layer k (shown on the x axis), we record the \% of outputs which report an answer which is neither the parametric nor the contextual answer (as decided by an LLM-as-a-Judge).}
  \label{fig:back-patch_neither}
\end{figure*}

\begin{figure*}[t]
  \includegraphics[width=\textwidth]{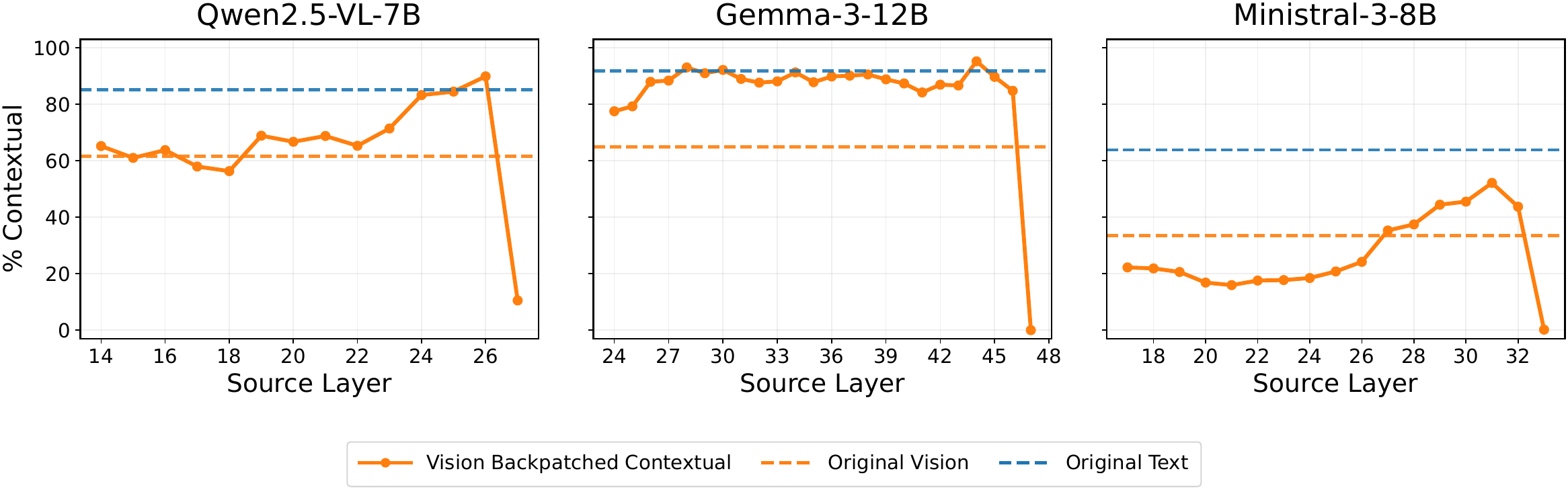}
  \caption{Contextual reporting rate under back-patching. After back-patching from source layer k (shown on the x axis), we record the \% of outputs which report an answer that matches the factoid in the context (as decided by an LLM-as-a-Judge).}
  \label{fig:back-patch_context}
\end{figure*}

\subsection{Back-patching}

While our back-patching experiments utilized the \% parametric metric, it is plausible that the decrease in parametric output is due to the model switching its answer to neither option, rather than switching to the contextual answer. As shown in Figure~\ref{fig:back-patch_neither}, the neither rate remains near 0 for all models until the last layer for Qwen and Gemma. For Ministral, the neither rate starts to climb around layer 30. While this suggests that very late representations are out-of-distribution or no longer relate to the original queried entity, we note that the back-patching layer 29 appears to lessen the gap. Furthermore, in Figure~\ref{fig:back-patch_context}, we see that the contextual rate for Ministral also steadily rises towards the text baseline. Therefore, our conclusion in Section~\ref{sec:mechanism:proc-time} remains that the behavioral divergence under conflict is associated with a lack of early alignment.

\section{Prompt-Based Mitigation Details}
\label{sec:prompting_details}

\begin{figure*}[t]
\centering

\begin{tcolorbox}[
    colback=white,
    colframe=black,
    boxrule=0.5pt,
    arc=1pt,
    left=5pt,
    right=5pt,
    top=5pt,
    bottom=5pt,
    width=0.98\linewidth
]

\scriptsize
\begin{verbatim}
Reason through this question by:
1. Determine the entity referenced in the question
2. Consider both the context information and your knowledge
3. Derive the answer to the question

Required JSON fields:
- entity: A string containing the name of the entity in 1-5 words
- reasoning: A string containing 3-4 sentences explaining your thought process
- answer: A string containing the final answer in 1-3 words

Example 1:

Input: Context information is below.
---------------------
Adobe was founded in 1982.
Adobe was founded by John Warnock and Charles Geschke.
Adobe is best known for manufacturing smartphone processors.
---------------------

Given the context information and your knowledge,
answer the question about the following entity.

Entity: Adobe.

Query: What product is this company most associated with?

Output:
{
  "entity": "Adobe",
  "reasoning": "The entity is Adobe. The founding year and
  founders match my background knowledge about the company.
  However, the statement that Adobe is known for manufacturing
  smartphone processors conflicts with widely known information
  about Adobe. Adobe is most strongly associated with creative
  software products.",
  "answer": "Photoshop"
}

Example 2:

Input: Context information is below.
---------------------
Nintendo was founded in 1889.
Nintendo was founded by Fusajiro Yamauchi.
Nintendo is best known for producing kitchen appliances.
---------------------

Given the context information and your knowledge,
answer the question about the following entity.

Entity: Nintendo.

Query: What product is this company most associated with?

Output:
{
  "entity": "Nintendo",
  "reasoning": "The entity is Nintendo. The founding year and
  founder match my background knowledge about the company,
  which makes the context appear internally grounded. Although
  my prior knowledge associates Nintendo with video games,
  the provided context identifies kitchen appliances as the
  product category associated with the company. Considering
  the context information and the question, I answer with the
  product category given in the context.",
  "answer": "Kitchen appliances"
}
\end{verbatim}

\end{tcolorbox}

\caption{Representative few-shot Chain-of-Thought (CoT) demonstrations used in our context--memory conflict prompting setup. The prompt instructs the model to explicitly identify the entity, reason about the relationship between contextual and parametric knowledge, and produce a structured JSON output containing the reasoning trace and final answer.}
\label{fig:cot_prompt_example}

\end{figure*}

In Section~\ref{sec:mitigation}, we investigate black-box mitigation methods. Specifically, we evaluate whether modality divergence can be reduced through chain-of-thought prompting (Figure~\ref{fig:cot_prompt_example}) or by presenting the conflicting contextual evidence visually (Figure~\ref{fig:visual_context_prompt_text_entity}).

\subsection{Chain-of-Thought Prompting}

For the CoT setting, we modify the standard prompt with explicit reasoning instructions and require the model to generate structured JSON outputs. Specifically, the model is instructed to:
\begin{enumerate}
    \item Identify the entity referenced in the image or text prompt.
    \item Consider both the supplied context information and its internal knowledge.
    \item Derive the final answer to the question.
\end{enumerate}
To encourage consistent formatting and controllable parsing, we require the model to produce outputs containing three fields:
\begin{itemize}
    \item \texttt{entity}: the resolved entity name,
    \item \texttt{reasoning}: a short 3--4 sentence explanation,
    \item \texttt{answer}: the final 1--3 word answer.
\end{itemize}
We additionally provide two in-context examples. To avoid the in-context examples biasing the model to either parametric or in-context reporting, we ensure the two examples include a parametric output and a contextual output. Furthermore, in-context examples discuss a domain in none of our datasets (companies) to minimize any biases that might arise from domain similarity.

\begin{figure}[t]
\centering

\begin{tcolorbox}[
    colback=white,
    colframe=black,
    boxrule=0.5pt,
    arc=1pt,
    left=5pt,
    right=5pt,
    top=5pt,
    bottom=5pt,
    width=0.98\linewidth
]

\scriptsize
\begin{verbatim}
Context information is below.
---------------------
The following image provides
the retrieved context entity:

<start_of_image>
<context_image>
<end_of_image>

The person pictured was born on
December 13, 1989.
The person pictured is American.
The person pictured is a novelist.
---------------------

Given the context information and your
knowledge, answer the question about
the following entity.

Entity: Taylor Swift.

Query: What is the occupation of the entity?
Answer with only 1-3 word(s).
\end{verbatim}

\end{tcolorbox}

\caption{Representative prompt template for the visual-context setting with a textual queried entity. The retrieved contextual entity is provided visually, while the queried entity is presented textually.}
\label{fig:visual_context_prompt_text_entity}

\end{figure}

\subsection{Visual Context Prompting}

Our second prompting approach replaces the textual entity reference in the retrieved context with an image of the retrieved entity. For example, instead of supplying the statement ``Taylor Swift is a novelist,'' we provide an image of Taylor Swift alongside the modified statement ``The person pictured is a novelist.'' This setup forces the model to resolve the retrieved contextual entity visually rather than textually. We evaluate both a same-image setting, where the identical image is reused across the context and query, and a different-image setting, where a distinct image of the same entity is provided in the context.

\section{Birth-Year Analysis}
\label{sec:alt_rel}

\begin{figure*}[t]
  \includegraphics[width=\textwidth]{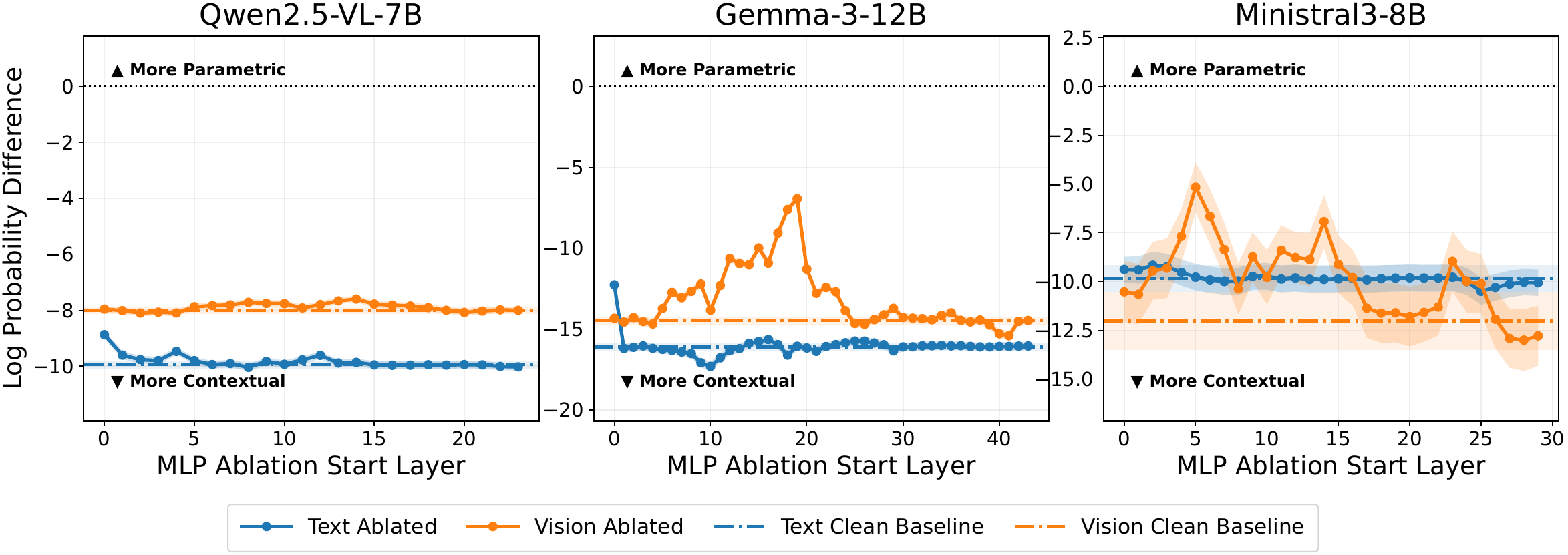}
  \caption{Behavioral impact of MLP ablation across entity positions for celebrity birth-years. }
  \label{fig:MLP_BY_Ablation}
\end{figure*}

Our mechanistic analysis focused on localizing and specifying the mechanism which leads to divergent behavior across modalities. Therefore, we restricted our mechanistic analysis to subcategories that resulted in the largest divergence. However, to understand whether suppression occurs when behavior is consistent in both modalities, we conduct a MLP ablation experiment on the birth year category. We chose birth-year as it has the most retained examples after filtering out of the subcategories with little to no divergence. In Figure~\ref{fig:MLP_BY_Ablation} we observe that ablating the MLPs in both modalities do not result in a shift towards the contextual answer, which implies that these MLPs likely do not result in parametric promotion. This could be because birth year information may be encoded differently (or not at all, since models are not performing well on this subcategory) and thus do not undergo the parametric promotion mechanism. 

\section{Visual Context Suppresses MLPs}
\label{sec:vis_ctxt_mech}

\begin{figure*}[t]
  \includegraphics[width=\textwidth]{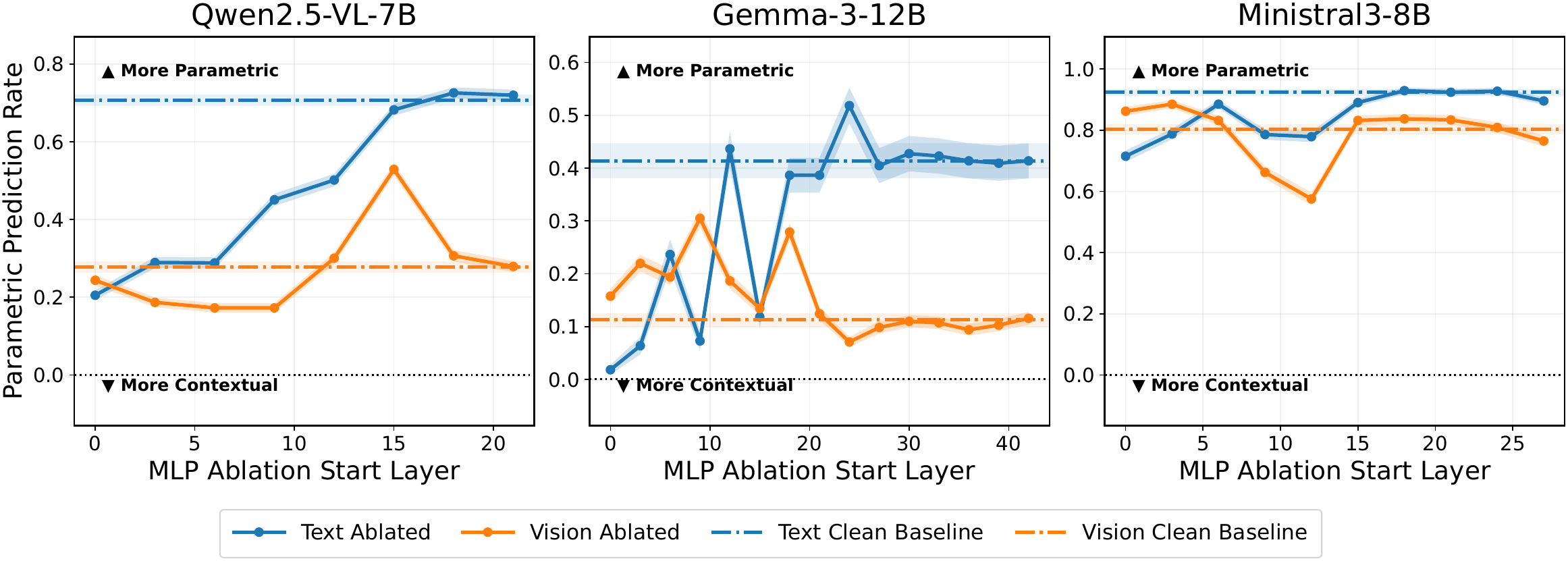}
  \caption{Behavioral impact of MLP ablation under visual context conflict. The x-axis indicates the starting layer of the ablated MLP span. The y-axis represents parametric reporting rate. We utilize a constant span of five layers for all the models. The shaded region represents $\pm$ one standard error of the mean.}
  \label{fig:vctxt_mlp_ablate}
\end{figure*}

In Section~\ref{sec:vctxt_med_gap}, we observe that providing a visual reference to the entity in the context resulted in an inverse of behavior. Such that visual entities now displayed a bias towards in-context information, and textual entities reported parametric knowledge at a higher rate. We therefore infer that entity alignment is a prerequisite for MLP suppression. To further validate this inference, we repeat the MLP ablation experiment from Section~\ref{sec:mechanism:mlps} under this setting. As shown in Figure~\ref{fig:vctxt_mlp_ablate}, ablating early-layer MLPs for textual entities in Qwen and Gemma results in a large shift towards the contextual answer, implying the MLPs promoted the parametric answer. This pattern is the inverse of that observed in Figure~\ref{fig:MLP_Ablate_Results}, suggesting that the MLPs are now suppressed for visual entities and, consequently, that early entity alignment mediates MLP suppression in these two models. For Ministral, we observe that ablating the early-layer MLPs still results in a shift towards the contextual answer for textual entities, implying that the textual MLPs are no longer suppressed. However, we observe a similar effect for visual entities, suggesting that their MLPs are also not fully suppressed. This may explain why providing visual context results in a relatively smaller reduction in the modality gap for Ministral compared to the other two models.

\section{Visual Inference of Relations Does Not Significantly Impact Results}
\label{sec:vis_inf}

\begin{figure*}[t]
  \includegraphics[width=\textwidth]{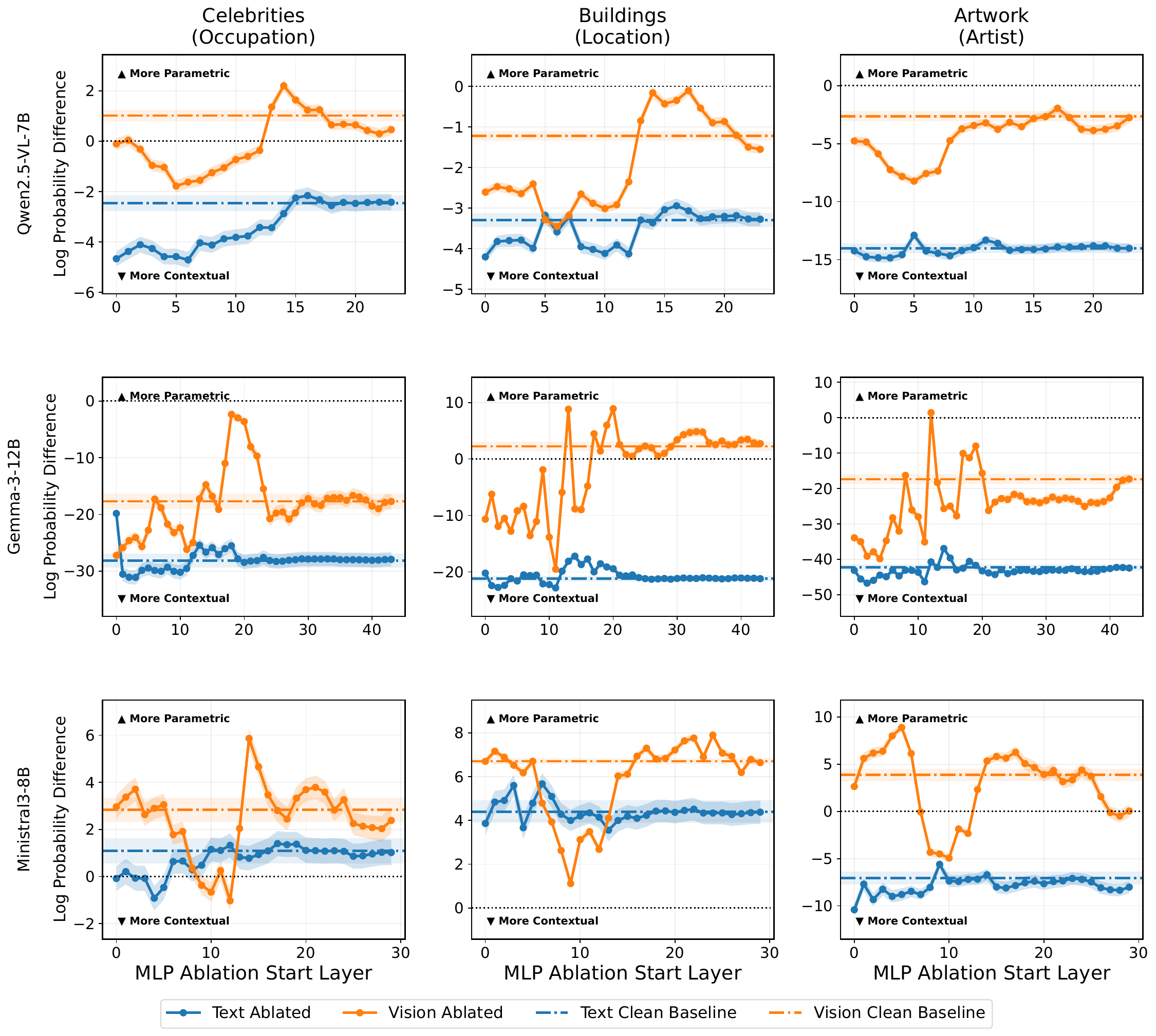}
  \caption{Behavioral impact of MLP ablation split by subcategories. The x-axis indicates the starting layer of the ablated MLP span. The y-axis represents the margin defined in Equation \ref{eq:margin}. We utilize a constant span of five layers for all the models. The shaded region represents $\pm$ one standard error of the mean.}
  \label{fig:sep_mlp_abl}
\end{figure*}

\begin{figure*}[t]
  \includegraphics[width=\textwidth]{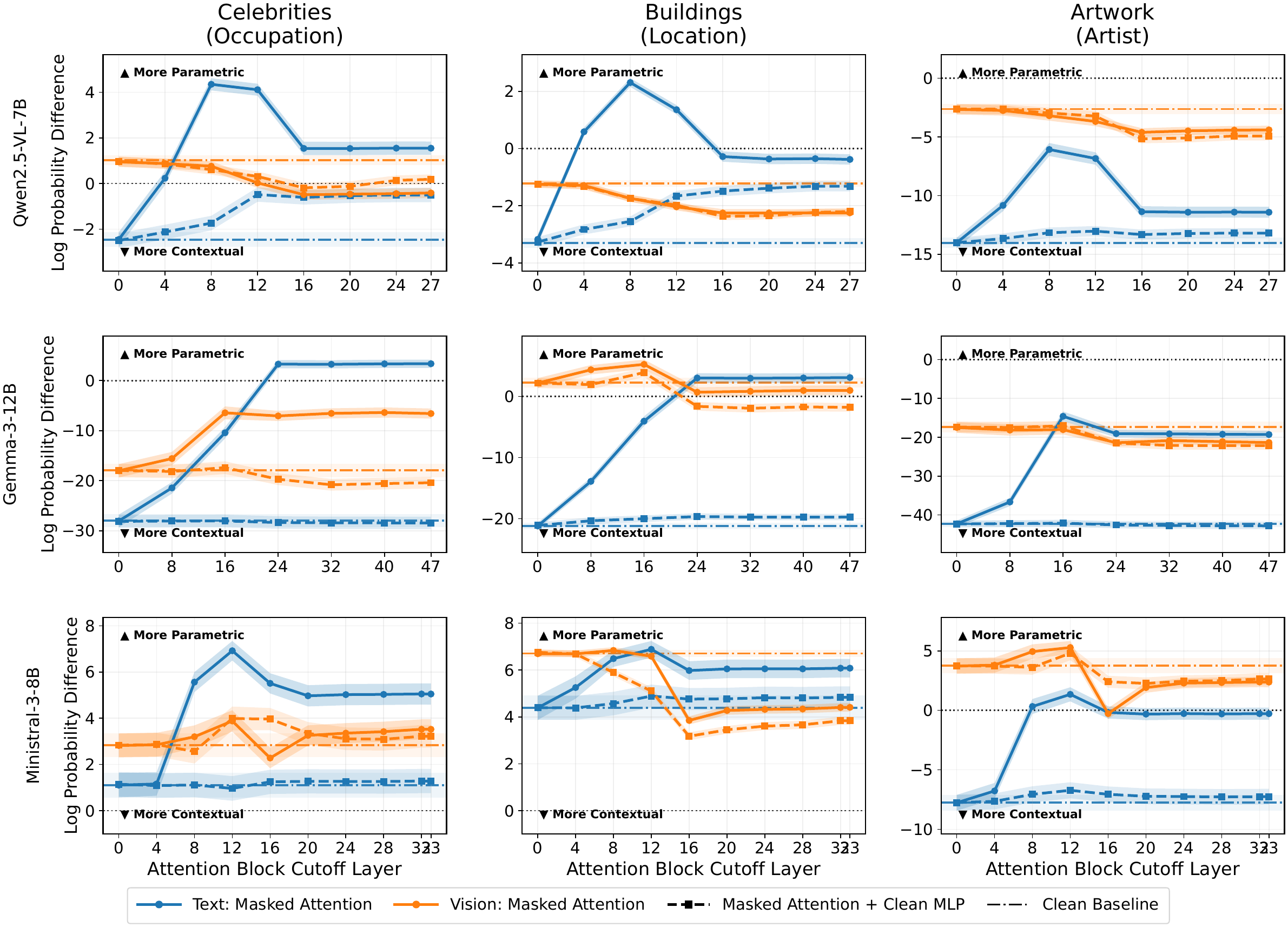}
  \caption{Effect on preference margin when masking attention from entity tokens to the retrieved context split by subcategories.}
  \label{fig:sep_Attn_Block_Results}
\end{figure*}

\begin{figure*}[t]
  \includegraphics[width=\textwidth]{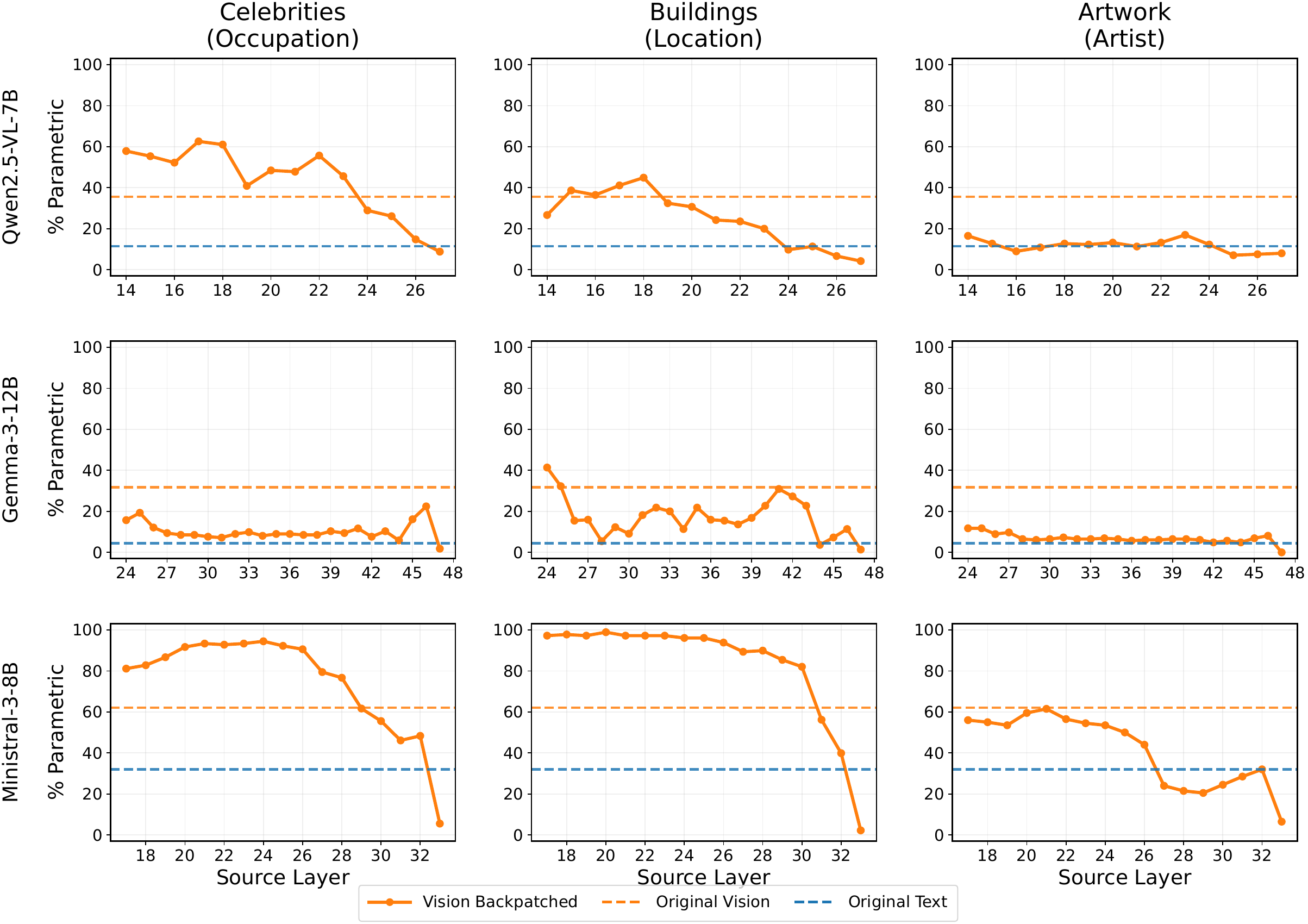}
  \caption{Parametric answer reporting rate under back-patching split by subcategories. We patch activations from the visual-token span of a specific source layer (ranging from layer $\frac{L}{2}$ to the final layer $L$, shown on the x-axis) into the first layer. The solid orange line tracks the model’s resulting parametric answer reporting rate (\%). For comparison, the dashed orange and dashed blue horizontal lines represent the baseline visual and textual parametric reporting rates, respectively.}
  \label{fig:sep_backpatch}
\end{figure*}

Within our dataset, a few examples within the celebrity category include images where the occupation can be visually inferred. For example a model could infer LeBron James is a basketball player due to the fact that he is wearing a basketball jersey. In this case, it may remain unclear whether the model undergoes parametric retrieval rather than visual inference. To assess whether our analysis was impacted, we split our mechanistic experiments by the three datasets, as building location and artwork artist are significantly harder to infer without processing the main subject.  In Figure~\ref{fig:sep_mlp_abl} we see that the MLP ablation experiment across all three subcategories results in strikingly similar trends. Figure~\ref{fig:sep_Attn_Block_Results} highlights the attention masking experiment split by dataset. We observe that for Qwen and Ministral, the trend is nearly identical for all three datasets. However, for Gemma, we note that the celebrity dataset seems to undergo a small amount of suppression, which may explain why the subcategory displayed the smallest modality gap. Nevertheless, the magnitude of this suppression remains substantially smaller than in the text condition and therefore does not meaningfully affect our analysis or conclusions. Finally, in Figure~\ref{fig:sep_backpatch}, we observe that although the precise layer at which the modality gap closes varies across datasets, the general trend of later-layer backpatching closing the gap is consistent across all three datasets.

\section{Gemma-3-27B Suppresses MLPs Across Both Modalities}
\label{sec:gemma_mech}
\begin{figure*}[t]
  \includegraphics[width=\textwidth]{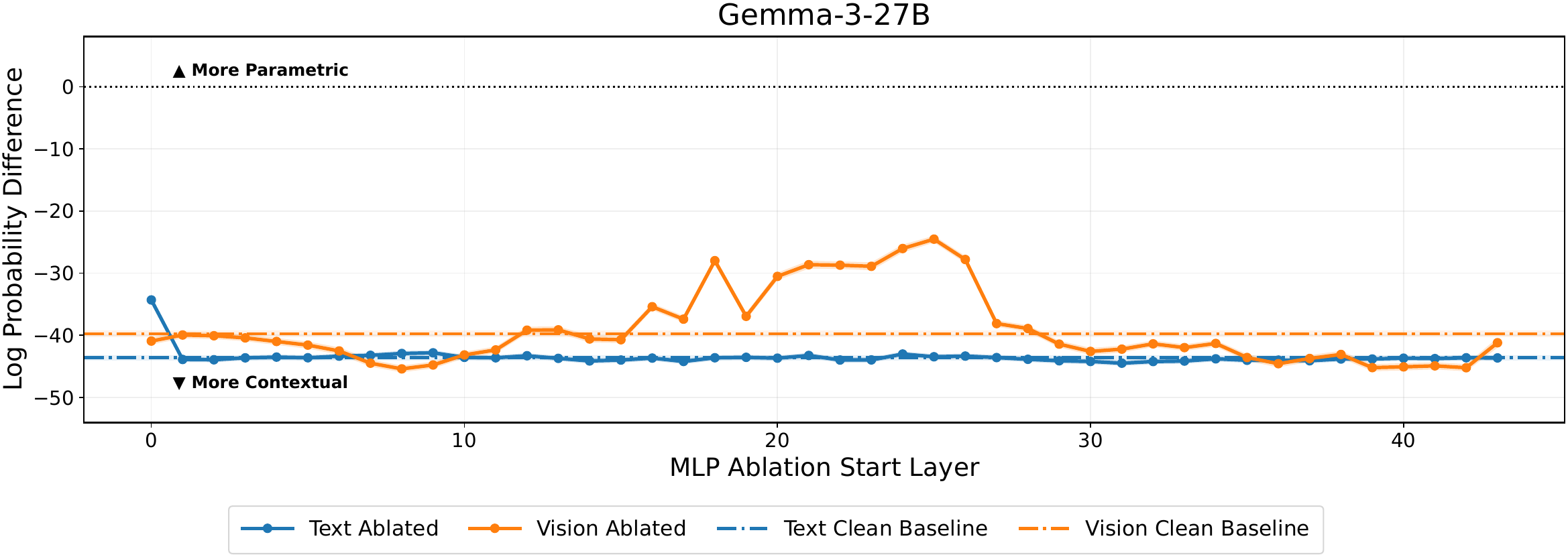}
  \caption{Behavioral impact of MLP ablation across entity positions for Gemma-3-27B. The x-axis indicates the starting layer of the ablated MLP span. The y-axis represents the margin defined in Equation \ref{eq:margin}. We utilize a constant span of five layers. The shaded region represents $\pm$ one standard error of the mean.}
  \label{fig:GEMMA_MLP_Ablate_Results}
\end{figure*}

\begin{figure*}[t]
  \includegraphics[width=\textwidth]{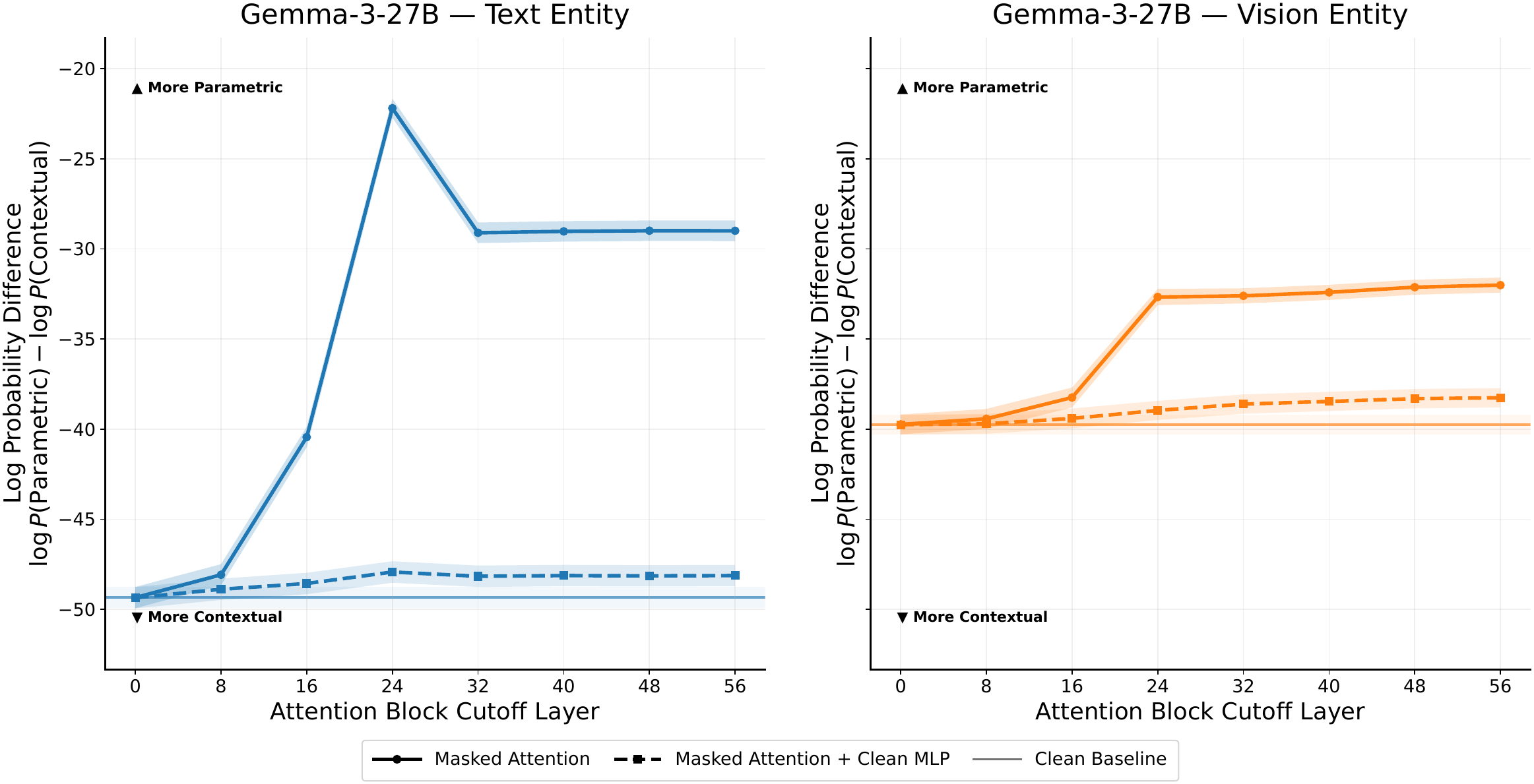}
  \caption{Effect on preference margin when masking attention from entity tokens to the retrieved context for Gemma-3-27B. For each cutoff layer $k$ on the x-axis, multi-head self-attention modules from layers $0$ through $k$ were masked so that entity-token positions could not attend to the retrieved context. Solid lines show the standard masking intervention, where downstream entity-token MLPs receive the attention-masked residual stream as input. Dashed lines show the control intervention, where the entity-token MLP outputs added to the residual stream are replaced with their corresponding outputs from the clean, unmasked run. Dash-dotted horizontal lines denote the clean-run margin for each modality. Shaded regions indicate $\pm$ one standard error of the mean (SEM).}
  \label{fig:GEMMA_Attn_Block_Results}
\end{figure*}

In Table~\ref{tab:parametric_by_relation}, we observe that all the models tested tend to show a modality gap in behavior except for Gemma-3-27B. To test whether this is due to symmetric MLP suppression, we run the MLP ablation experiment from Section~\ref{sec:mechanism:mlps}. In Figure~\ref{fig:GEMMA_MLP_Ablate_Results}, we observe that ablating the MLPs over the entity tokens results in a much smaller shift toward the contextual answer. This implies that the MLPs no longer promote the parametric answer. To verify that this is mediated through the attention heads attending to the conflict, we conduct the attention masking experiment from Section~\ref{sec:mechanism:attn}. As shown in Figure~\ref{fig:GEMMA_Attn_Block_Results}, we see that parametric promotion is suppressed under conflict for both modalities, as both shift towards the parametric output under masking. This emphasizes that the symmetric suppression across modalities mediates a reduced gap. 

\end{document}